\documentclass[journal,transmag]{IEEEtran}

\ifCLASSINFOpdf

\else

\fi

\usepackage{color}

\usepackage{CJK}
\usepackage{amsmath}
\usepackage{amssymb}
\usepackage{ctable} 
\usepackage{tabularx}
\usepackage{setspace}
\usepackage{comment}
\usepackage[numbers]{natbib}

\usepackage{graphicx}
\usepackage{subfigure}
\usepackage{multirow}
\usepackage{booktabs}
\usepackage{ctable}
\usepackage{ctable}
\usepackage{makecell}

\usepackage{floatrow}
\newfloatcommand{capbtabbox}{table}[][\FBwidth]

\usepackage{bbm}
\usepackage{dsfont}

\usepackage{algorithm,algorithmic}
\usepackage{multicol}

\begin{document}

\bibliographystyle{IEEEtran}

\title{Distribution Learning Based on Evolutionary Algorithm Assisted Deep Neural Networks for Imbalanced Image Classification\\}

\author{\IEEEauthorblockN{Yudi Zhao\IEEEauthorrefmark{1},
Kuangrong Hao\IEEEauthorrefmark{1,*},
Chaochen Gu\IEEEauthorrefmark{2}, 
Bing Wei\IEEEauthorrefmark{1}
}
\IEEEauthorblockA{\IEEEauthorrefmark{1}College of Information Sciences and Technology, Donghua University, Shanghai 201620, P. R. China}
\IEEEauthorblockA{\IEEEauthorrefmark{2}Department of Automation, Shanghai Jiao Tong University, Shanghai 200240, P. R. China}

\thanks{*Corresponding author: Kuangrong Hao (email: krhao@dhu.edu.cn)}}

\markboth{Journal of \LaTeX\ Class Files,~Vol.~14, No.~8, August~2015}%
{Shell \MakeLowercase{\textit{et al.}}: Bare Demo of IEEEtran.cls for IEEE Transactions on Magnetics Journals}

\IEEEtitleabstractindextext{%
\begin{abstract}
To address the trade-off problem of quality-diversity for the generated images in imbalanced classification tasks, we research on over-sampling based methods at the feature level instead of the data level and focus on searching the latent feature space for optimal distributions.
On this basis, we propose an i\textbf{M}proved \textbf{E}stimation \textbf{D}istribution \textbf{A}lgorithm based \textbf{L}atent feat\textbf{U}re \textbf{D}istribution \textbf{E}volution (MEDA\_LUDE) algorithm, where a joint learning procedure is programmed to make the latent features both optimized and evolved by the deep neural networks and the evolutionary algorithm, respectively.
We explore the effect of the Large-margin Gaussian Mixture (L-GM) loss function on distribution learning and design a specialized fitness function based on the similarities among samples to increase diversity.  
Extensive experiments on benchmark based imbalanced datasets validate the effectiveness of our proposed algorithm, which can generate images with both quality and diversity. 
Furthermore, the MEDA\_LUDE algorithm is also applied to the industrial field and successfully alleviates the imbalanced issue in fabric defect classification.

\end{abstract}

}

\maketitle

\IEEEdisplaynontitleabstractindextext

\IEEEpeerreviewmaketitle

\section{Introduction}
\IEEEPARstart{I}{mbalanced} 
problems have always occurred in real-world collected data, where the high degree of variation is exhibited in the number of different categories due to data characteristics, problems in the collection process, or human factor.
Traditional classification methods are built on the premise of the data balance and hence easily confront a sharp drop in classification performance when handling imbalanced data with highly skewed distributions \cite{zhang2016transfer}.
Since imbalanced data is widespread in people's lives and industrial production, such as credit card fraud recognition, industrial fault detection, and diagnosis, as well as cancer diagnosis and treatment, imbalanced classification problems have become challenges in the fields of machine learning and data mining \cite{8528854}.

Traditional over-sampling based techniques to address imbalanced problems mainly synthesize samples randomly among the minority classes \cite{rosales2022handling}.
Random over-sampling (ROS) \cite{liu2007generative} is the simplest way to increase the ratio of minority samples, where the valuable information is not increased as the sample count.
To improve the generalization ability, Chawa et al.  \cite{chawla2002smote} designed a synthetic minority over-sampling technique (SMOTE) based on a neighbor strategy among the minority samples. However, noise could be introduced to the synthesis process, and the problem of distribution marginalization probably happens, especially when noisy samples appear at the boundary of the positive samples and the negative samples.
For this, borderline-SMOTE \cite{han2005borderline} is further raised to operate synthesis around the boundaries.
He et al. \cite{he2008adasyn} proposed an adaptive synthetic (ADASYN) sampling method that adaptively synthesizes minority samples according to the distribution as well as the learning difficulty of the minority samples.
Besides, the majority weighted minority oversampling technique (MWMOTE) \cite{barua2012mwmote} innovatively assigns different weights to the hard-to-learn minority samples. The weight is determined by their distances to the nearest majority samples.
However, traditional over-sampling based methods exist limitations in classification tasks when dealing with high-dimensional images. 
Due to the powerful learning ability, deep learning based over-sampling techniques have shown superiority in imbalanced problems.
Variational autoencoder (VAE), as a generative model,  has strong capacity for image generation due to its variational inference ability between the complex data and the low-dimensional latent feature space \cite{9216561}.
Wan et al. \cite{wan2017variational} utilized VAE to generate the minority images and successfully solved the imbalanced problem in image classification tasks.
Dai et al. \cite{dai2019generative} further leveraged the information from the majority samples based on the VAE, which achieved great success in the medical field.
Moreover, the conditional VAE (CVAE) \cite{lim2018molecular} is a modification of VAE to guide and control the image synthesis process by incorporating prior constraints.

Even though deep learning based over-sampling models have made achievements in imbalanced classification problems, the quality and diversity of samples can not always be guaranteed simultaneously, especially when the existing samples are limited. To tackle this puzzle, some recent works explore how to obtain superior quality and diversity trade-off in generation tasks. 
Yang et al. \cite{yang2020generative} improved synthesis quality by removing detrimental ones based on influence functions and maximized the diversity by picking out compelling examples from the generated pool for the application of commonsense reasoning. Still, both of the selections are based on the level of samples, which is not so efficient when requiring a large amount of generative data. 
Du et al. \cite{9387399} proposed a multiconstraint generative adversarial network (MCGAN) to ensure the similarity, diversity, and correct category of the synthetic aperture radar (SAR) images in the field of automatic target recognition (ATR), where the diversity 
is increased by adding Gaussian noise to the input vector of the generator. Similarly, noise perturbation is also employed as well as infinite generative samples to enhance the discrimination, which can boost the high-quality and diverse generation \cite{yang2021data}. However, noise-based methods are accompanied by randomness and are less controllable.
In \cite{costa2020exploring}, a new quality-diversity based evolutionary algorithm is applied to GANs to upgrade the exploration of search space and provide better solutions for generators and discriminators. Hence, the final generators competitively synthesized high-quality samples. Although this method can be trained controllably with objectives, it aims at guiding the evolution of GANs to discover more efficient models rather than data distribution.

The synthesis of data depends on the generator and its input vectors sampled from the latent distributions. Meanwhile, it has proved that the closer the generated distribution is to the real distribution, the higher the quality and diversity of the synthesized data achieve \cite{alihosseini2019jointly, li2020relation}.
Therefore, in imbalanced problems, distribution learning is essential for deep learning based oversampling methods for image generations. 
However, existing deep learning based methods are prone to the risk of overfitting, and the learned distribution is likely restricted within a limited range far from the real distribution when given training data is imbalanced or even insufficient.
This paper employs an evolutionary algorithm to learn distributions to alleviate this problem. 
A similar idea defined as latent variable evolution (LVE) is first introduced for dictionary attack \cite{bontrager2018deepmasterprints}. This research work adopts the covariance matrix adaptation evolution strategy (CMA-ES) to evolve the input variables of GANs for fingerprint synthesis. However, their objective is to match the impostors with as many subjects in the real fingerprint images as possible during the evaluation of the recognition system. 
Besides, LVE is also applied to procedural content generation (PCG) of video game levels
 \cite{volz2018evolving, 
giacomello2019searching, thakkar2019autoencoder}.
Volz et al. \cite{volz2018evolving} generated playable Mario levels by applying CMA-ES to GANs.
The same approach is also exploited to search the latent space for producing DOOM levels with an excellent grade of novelty and variety \cite{giacomello2019searching}. 
In addition, Thakkar et al.\cite{thakkar2019autoencoder} proposed to merge a multi-population evolutionary algorithm into a multi-channel autoencoder for Lode Runner level generations and realized the goal of playability and connectivity. 
However, these approaches are not applicable to image generation for imbalanced problems
since the fitness functions are devised according to the desired properties of different video games rather than objectives like quality or diversity.

In this paper, we assume that the latent features follow a multivariate Gaussian mixture (GM) distribution 
and apply the estimation of distribution algorithm (EDA) \cite{zhou2007survey} to evolve the latent feature distribution, which is simultaneously optimized by deep neural networks with the large-margin Gaussian Mixture (L-GM) loss for quality-diversity trade-off. 
Considering we aimed at optimizing and searching for the optimal distribution in the latent space while VAEs already exist a prior probability distribution for latent variables, thus the autoencoders (AEs) are applied where
the input images are mapped to the feature space by an encoder, and then the latent features are decoded to the synthesized samples. 
Meanwhile, two classifiers, including a latent feature classifier and an image classifier, are employed to control and improve the feature distribution learning process for better diversity, and train the generator for better quality. 
Our contributions are summarized as follows:
(1) A deep neural network based architecture for imbalanced problems is devised, where an iMproved EDA (MEDA) is utilized to evolve the latent feature distributions, and the proposed method can well balance the quality and diversity of generations.
(2) A specialized fitness function is designed for the MEDA according to the similarities among samples. The fitness function will guide the search for the latent variables to enhance the diversity.
(3) We exploit the function of L-GM loss under a more complex assumption, where the covariance matrix is assumed to be a variable participating in the optimization of deep neural networks and the evolution of MEDA instead of being an identity matrix. 
(4) Four training phases are innovatively programmed in the MEDA\_LUDE algorithm, which can achieve excellent performance on benchmark datasets and be successfully applied to the industrial field.

The remainder of the paper is organized as follows: Section \uppercase\expandafter{\romannumeral2} introduces concepts of EDA and L-GM loss. Detailed descriptions of our proposed method are given in \uppercase\expandafter{\romannumeral3}. In \uppercase\expandafter{\romannumeral4}, experimental results and analysis are displayed. Finally, we present our conclusions.  

\section{Background}

\subsection{Estimation of Distribution Algorithms}
Estimation of Distribution Algorithms (EDAs) are a novel branch of statistical-model-based evolutionary algorithms and provide a microscopical paradigm where the population evolves by learning from the probabilistic distribution model \cite{yang2016multimodal}. 
Although EDAs have developed various implementations, the core of them can be summed up as two main steps \cite{zhou2007survey, liang2018enhancing}:
(1) Construct the probabilistic model of the solution space and select the superior individuals based on the evaluation. The new explicit probabilistic model can describe the distribution of the updated solution space. (2) Generate a new population by sampling randomly from the current probabilistic model, and then repeat steps (1)-(2) until the termination criterion is met.
 
In this paper, we assume that the latent features follow a multivariate Gaussian mixture model (GMM). Hence, we give a brief introduction of the basic Gaussian distribution based EDA \cite{xue2017Swarm}. 
Assume that $N$ represents the population size, $k$ and $Iter$ are the iterations and the maximum number of the iterations, respectively. $\eta$ denotes the sampling rate of the superior population.
The general procedure is as follows:
\begin{enumerate}
\item Randomly generate $N$ initial solutions as a population, which is denoted as $pop(0)=\{x^{1}, x^{2},\cdots, x^{N}\}$, here $k=0$.
\item Calculate the fitness of each individual in $pop(k)$.
\item Rank the individuals in a descending order based 	on fitnesses, and keep the former $\eta N$ individuals as the superior population $spop(k)$.
\item if $k=Iter$, terminate the algorithm; Otherwise, continue step 5).
\item Calculate the mean and the variance of $pop(k)$ and$spop(k)$ for each dimension.
\item Update the mean and the variance according to rules below:
\begin{align}
\mu=\gamma\mu_{new} + (1-\gamma)\mu \nonumber\\
\sigma^{2}=\gamma\sigma^{2}_{new} + (1-\gamma)\sigma^{2} \nonumber
\end{align}
where $\mu$, $\sigma^{2}$ denote the means and the variances  of $pop(k)$, while $\mu_{new}$, $\sigma^{2}_{new}$ represent those of $spop(k)$. $\gamma$ is the weighting coefficients.
\item Sample from the Gaussian distribution with the current means and variances, then generate a new population.
\item $k:=k+1$, and go to step 2).
\end{enumerate} 

In our MEDA\_LUDE algorithm, the basic EDA mentioned above is improved and adopted to evolve the latent feature distribution for further quality and diversity enhancement in data generation.

\subsection{Large-margin Gaussian Mixture Loss}
The L-GM loss is proposed for deep neural networks in classification problems, where the extracted features are assumed to follow a GM distribution, and each category belongs to one ingredient \cite{wan2018rethinking}. 
The proposed L-GM loss is composed of a classification loss with large-margin and a likelihood regularization. The classification loss can enhance the discrimination and improve the generalization capability, and the likelihood regularization can drive the features to obey the GM distribution.  
Assume that $\mu_{k}$ and $\Sigma_{k}$ are the mean and covariance of a Gaussian distribution representing class $k$ in feature space, including total $K$ classes, and $x_{i}$ and $z_{i}$ denotes the extracted feature and its label of the $i-th$ sample. Then the L-GM loss \cite{wan2018rethinking} is defined as Eq. (1).
\begin{equation}
\begin{aligned}
\mathcal{L}_{GM,i}^{m}=&-log\frac{\vert\Sigma_{z_{i}}\vert^{-\frac{1}{2}}e^{-d_{z_{i}}(1+\alpha)}}{\sum_{k}\vert\Sigma_{k}\vert^{-\frac{1}{2}}e^{-d_{k}(1+\mathds{1}(k=z_{i})\alpha)}} \\
&+\lambda(d_{z_{i}}+\frac{1}{2}log\vert\Sigma_{z_{i}}\vert)
\end{aligned}
\end{equation}
\begin{equation}
\begin{aligned}
d_{k}=\frac{1}{2}(x_{i}-\mu_{k})^{T}\Sigma_{k}^{-1}(x_{i}-\mu_{k})  \quad k\in[1,K]
\end{aligned}
\end{equation}
where the first term of Eq. (1) is the classification loss, and the second term is the likelihood regularization. $\alpha$ is a non-negative parameter controlling the classification margin size $m$, and
the indicator function $\mathds{1}()$ equals $1$ if $k=z_{i}$, otherwise equals $0$. $\lambda$ is a non-negative weighting coefficient of the second term.

For simplicity, the authors hypothesize the covariance satisfies the hypothesis of the identity matrix, namely $\Sigma_{k}=I$. Hence the L-GM loss is finally simplified as below:
\begin{equation}
\begin{aligned}
\mathcal{L}_{GM,i}^{m}=-log\frac{e^{-d_{z_{i}}(1+\alpha)}}{\sum_{k}e^{-d_{k}(1+\mathds{1}(k=z_{i})\alpha)}}
+\lambda d_{z_{i}}
\end{aligned}
\end{equation}
\begin{equation}
\begin{aligned}
d_{k}=\frac{1}{2}(x_{i}-\mu_{k})^{2} \qquad k\in[1,K]
\end{aligned}
\end{equation}

In the proposed MEDA\_LUDE algorithm, L-GM loss is adopted to provide the means and covariances of GM distributions with great initial values for MEDA evolving.
Unlike the covariance assumption given above, we assume it to be a variable that engages in the optimization of deep neural networks and the evolution of evolutionary algorithms.
Therefore, the L-GM loss needs to be re-derived, and 
we should further adjust the implementations in a more complex situation.

\section{Methodology}
The proposed MEDA\_LUDE model consists of an encoder, a decoder, a latent feature classifier, and an image classifier, as shown in Fig. 1. 
The latent features learned from the encoder are supposed to follow a multivariate GM distribution, a mixture of  $K$ normal distributions, where $K$ represents the number of categories.
To reach the goal of quality-diversity trade-off in image generation for imbalanced classification, we program a training procedure through four phases based on the MEDA\_LUDE architecture, as depicted in Figs. 2-5, respectively.
\begin{figure}[htbp]
  \centering 
\includegraphics[width=0.95\linewidth]{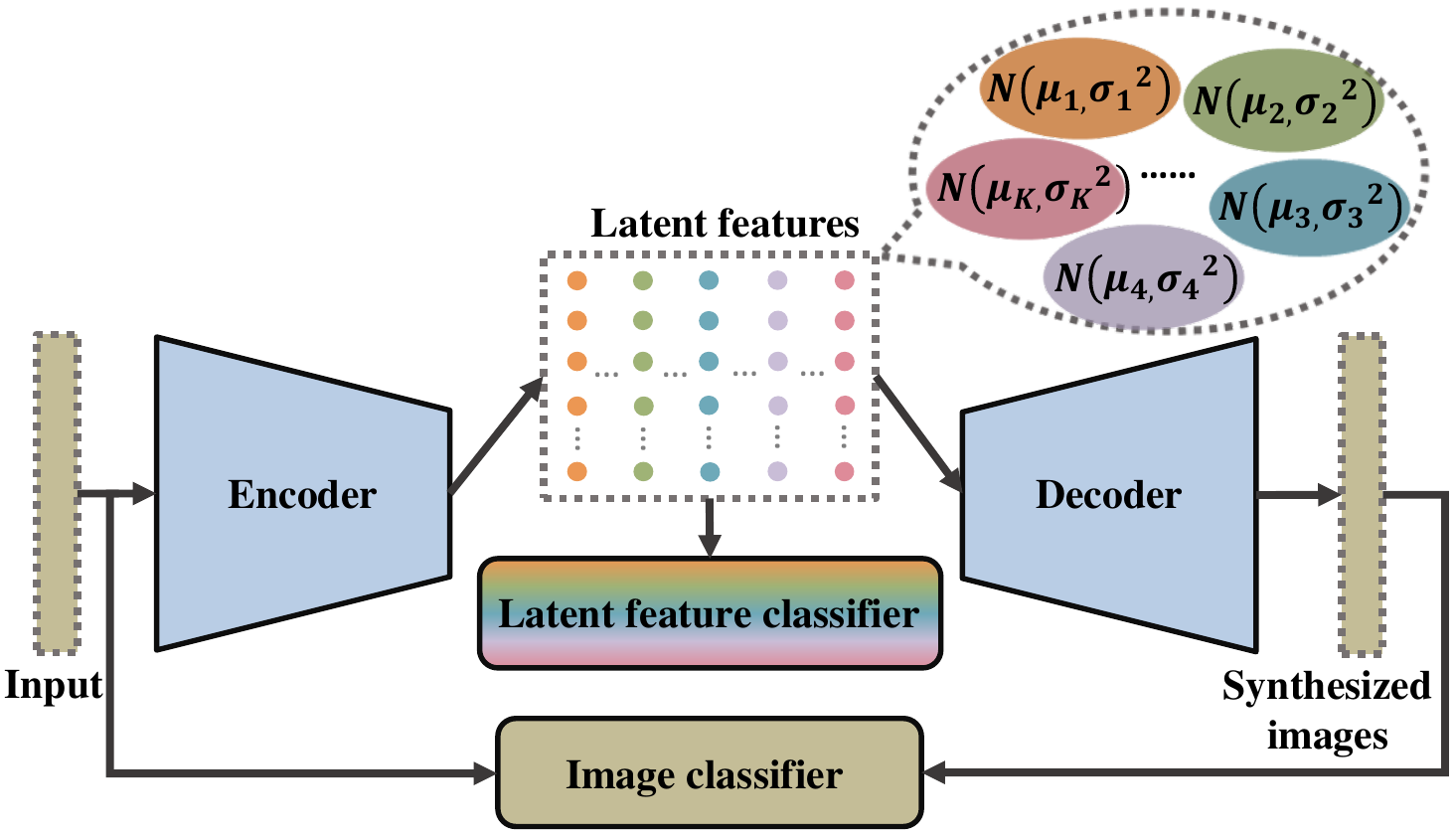}\\
  \caption{The MEDA\_LUDE architecture.}
  \vspace{-0.2cm}
\end{figure}

\subsection{Phase 1: Pre-training of MEDA\_LUDE}
\begin{figure}[htbp]
  \centering 
  \includegraphics[width=0.95\linewidth]{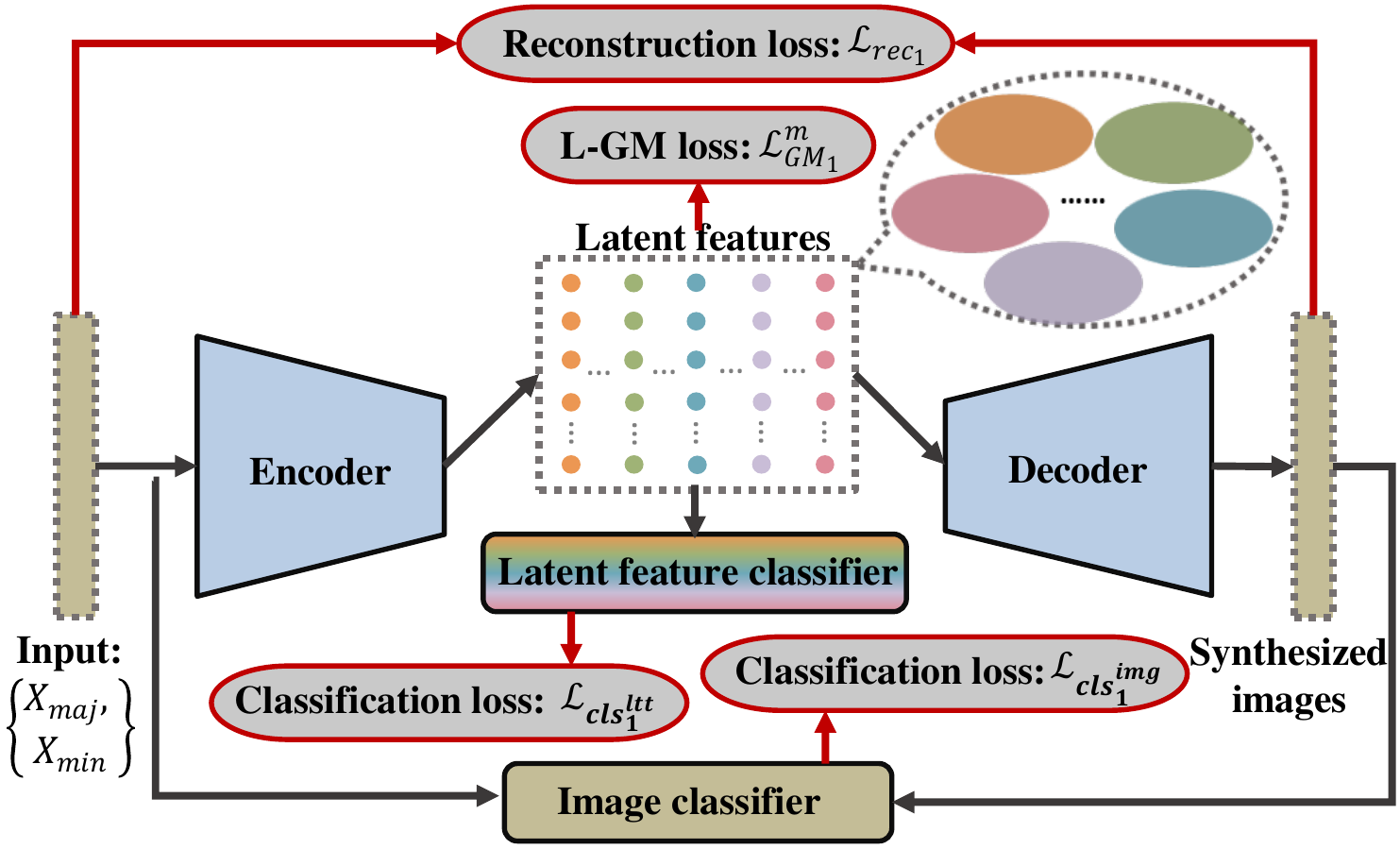}\\
  \caption{Phase 1: the parameters of the encoder, the decoder, the latent feature classifier, the image classifier, and the multivariate GM distributed latent features all participates in optimization.}
  \label{Fig. 2}
  \vspace{-0.2cm}
\end{figure}

As shown in Fig. 2, the input is composed of the majority samples $X_{maj}$ and the minority samples $X_{min}$.
The encoder, the decoder, and the two classifiers are pre-trained with reconstruction loss and classification loss. Meanwhile, the L-GM loss is also brought to the total loss, driving the latent features to follow a multivariate GM distribution automatically and giving a wise initialization for evolution operations without prior knowledge. 

Different from \cite{wan2018rethinking}, both the means and covariances of the multivariate GM distribution are hypothesized to be variables to be optimized in the MEDA\_LUDE algorithm. Here the assumption is given that $feat_{i}\in\mathds{R}^{1\times h}$ is the latent feature learned from the $i-th$ input image, where $h$ is the feature dimension. Correspondingly, the multivariate mean of class $k$ has the same dimension, denoted as $\mu_{k}\in\mathds{R}^{1\times h}$, meanwhile the covariance matrix $\Sigma_{k}\in\mathds{R}^{h\times h}$ is hypothesized to be diagonal, signified by $\Lambda_{k}\in\mathds{R}^{h\times h}$. The L-GM loss can be written as:
\begin{equation}
\begin{aligned}
\mathcal{L}_{GM,i}^{m}=&-log\frac{\vert\Lambda_{z_{i}}\vert^{-\frac{1}{2}}e^{-d_{z_{i}}(1+\alpha)}}{\sum_{k}\vert\Lambda_{k}\vert^{-\frac{1}{2}}e^{-d_{k}(1+\mathds{1}(k=z_{i})\alpha)}} \\
&+\lambda(d_{z_{i}}+\frac{1}{2}log\vert\Lambda_{z_{i}}\vert)
\end{aligned}
\end{equation}
\begin{equation}
\begin{aligned}
d_{k}=\frac{1}{2}(feat_{i}-\mu_{k})^{T}\Lambda_{k}^{-1}(feat_{i}-\mu_{k})  \quad k\in[1,K]
\end{aligned}
\end{equation}
where the first term to the right of the equal sign in Eq. (5) is the classification loss $\mathcal{L}_{cls,i}$, while the second one is the likelihood regularization $\mathcal{L}_{lkd,i}$.
In Eq. (6), $d_{k}$ calculates the squared Mahalanobis distance between the $i-th$ feature and the class $k$. 

With the non-ignorable and variable covariances $\Lambda_{k}$ in Eq. (6), the formulation needs to be re-derived when calculating the distance $\mathcal{D}$ between a batch of features and all the $K$ classes.
First, a batch of features are assumed as $FEAT=[feat_{1};feat_{2};\ldots;feat_{n}]\in\mathds{R}^{n\times h}$, where $n$ is the batch size. The multivariate mean with $K$ classes is
$M=[\mu_{1}; \mu_{2};\ldots; \mu_{K}]\in\mathds{R}^{K\times h}$. For convenience, the diagonal values of $\Lambda_{k}$ are extracted as a vector $\sigma_{k}^{2}=[\lambda_{k1},\lambda_{k2},\ldots,\lambda_{kh}]\in\mathds{R}^{1\times h}$, hence, the multivariate covariance is $\Sigma=[\sigma_{1}^{2};\sigma_{2}^{2};\ldots;\sigma_{K}^{2}]\in\mathds{R}^{K\times h}$.
Then the distance $\mathcal{D}$ is derived as follows:
\begin{equation}
\begin{aligned}
\mathcal{D}=&\frac{1}{2}\lbrace(FEAT\odot FEAT)\times \Sigma^{T}\\
&-2FEAT\times(M^{T}\odot \Sigma^{T})
+G
\end{aligned}
\end{equation}
\begin{equation}
G=
\begin{bmatrix}
g'\\g'\\\vdots\\g'
\end{bmatrix}_{n\times K}
\end{equation}
\begin{equation}
\begin{aligned}
g'=\sum_{j=0}^{h-1}g[j] \quad g'\in\mathds{R}^{1\times K}
\end{aligned}
\end{equation} 
\begin{equation}
\begin{aligned}
g=M^{T}\odot M^{T}\odot \Sigma^{T}\quad g\in\mathds{R}^{h\times K}
\end{aligned}
\end{equation} 
where `$\odot$' and `$\times$' denote the Hadamard product and the matrix multiplication, respectively. $j$ represents the row index of matrix $g$. The final distance $\mathcal{D}$'s dimension is $n\times K$, recording the squared Mahalanobis distance between $n$ features and $K$ classes.

To calculate the classification loss $\mathcal{L}_{cls}$, we assume $\vert\Lambda_{k}\vert^{-\frac{1}{2}}=q_{k}=e^{logq_{k}}$, and the logits $logit$ of $n$ samples in $K$ class can be calculated by:
\begin{equation}
\begin{aligned}
logit=-\mathcal{D}\odot(Ones+\alpha*label)+Q
\end{aligned}
\end{equation} 
\begin{equation}
\begin{aligned}
Q=
\begin{bmatrix}
logq\\logq\\\vdots\\logq
\end{bmatrix}_{n\times K}
\end{aligned}
\end{equation} 
\begin{equation}
\begin{aligned}
logq=[logq_{1},logq_{2},\ldots,logq_{K}]\in\mathcal{R}^{1\times K}
\end{aligned}
\end{equation} 
where $Ones$ is an $n\times K$ matrix of which each element is $1$.
$label$ is a  one-hot form from the labels of $n$ samples, whose dimension is $n\times K$.

Based on the $logit$, the classification $\mathcal{L}_{cls}$ of $n$ samples can be finally obtained by using 
cross entropy loss.

According to the previous assumption about $\vert\Lambda_{k}\vert^{-\frac{1}{2}}$, the likelihood regularization $\mathcal{L}_{lkd}$ can be derived as follows:
\begin{equation}
\begin{aligned}
\mathcal{L}_{lkd}=(\mathcal{D}-Q)\odot label
\end{aligned}
\end{equation} 

Finally, the L-GM loss can be achieved by:
\begin{equation}
\begin{aligned}
\mathcal{L}_{GM}^{m}=\mathcal{L}_{cls}+\lambda\mathcal{L}_{lkd}
\end{aligned}
\end{equation}

Finally, the phase loss $\mathcal{L}_{ph1}$ is defined by:
\begin{equation}
\begin{aligned}
\mathcal{L}_{ph_{1}}=&\beta_{rec_{1}}*\mathcal{L}_{rec_{1}}+\beta_{GM_{1}}*\mathcal{L}_{GM_{1}}^{m}\\
&+\beta_{cls^{ltt}_{1}}*\mathcal{L}_{cls^{ltt}_{1}}+
\beta_{cls^{img}_{1}}*\mathcal{L}_{cls^{img}_{1}}
\end{aligned}
\end{equation} 
\begin{equation}
\begin{aligned}
\mathcal{L}_{rec_{1}}=\mathcal{L}_{rec_{1}\_min}+\xi_{rec_{1}}*\mathcal{L}_{rec_{1}\_maj}
\end{aligned}
\end{equation} 
\begin{equation}
\begin{aligned}
\mathcal{L}_{GM_{1}}^{m}=\mathcal{L}_{GM_{1}\_min}^{m}+\xi_{GM_{1}}*\mathcal{L}_{GM_{1}\_maj}
\end{aligned}
\end{equation} 
\begin{equation}
\begin{aligned}
\mathcal{L}_{cls^{ltt}_{1}}=\mathcal{L}_{cls^{ltt}_{1}\_min}+\xi_{cls^{ltt}_{1}}*\mathcal{L}_{cls^{ltt}_{1}\_maj}
\end{aligned}
\end{equation} 
\begin{equation}
\begin{aligned}
\mathcal{L}_{cls^{img}_{1}}=&\mathcal{L}_{cls^{ori}_{1}\_min}+\xi_{cls^{ori}_{1}}*\mathcal{L}_{cls^{ori}_{1}\_maj}\\
&\mathcal{L}_{cls^{syn}_{1}\_min}+\xi_{cls^{syn}_{1}}*\mathcal{L}_{cls^{syn}_{1}\_maj}
\end{aligned}
\end{equation} 
where $\mathcal{L}_{rec_{1}}$, $\mathcal{L}_{GM_{1}}^{m}$, $\mathcal{L}_{cls^{ltt}_{1}}$, and $\mathcal{L}_{cls^{img}_{1}}$ are the reconstruction loss, the L-GM loss, the classification loss from the latent features and the images of Phase 1, respectively.
Meanwhile, $\beta_{rec_{1}}$, $\beta_{GM_{1}}$, $\beta_{cls^{ltt}_{1}}$, and $\beta_{cls^{img}_{1}}$ are their corresponding weighting coefficients.
Besides, $\mathcal{L}_{rec_{1}}$, $\mathcal{L}_{GM_{1}}^{m}$, and $\mathcal{L}_{cls^{ltt}_{1}}$ are all composed of the minority part and the majority part, with respective coefficients $\xi_{rec_{1}}$, $\xi_{GM_{1}}$, and $\xi_{cls^{ltt}_{1}}$ to alleviate the bias caused by imbalanced data.
Furthermore, $\mathcal{L}_{cls^{img}_{1}}$ includes the classification loss from both the original input image and the synthesized images, namely `$\mathcal{L}_{cls^{ori}_{1}\_min}+\xi_{cls^{ori}_{1}}*\mathcal{L}_{cls^{ori}_{1}\_maj}$' and `$\mathcal{L}_{cls^{syn}_{1}\_min}+\xi_{cls^{syn}_{1}}*\mathcal{L}_{cls^{syn}_{1}\_maj}$', where $\xi_{cls^{ori}_{1}}$ and $\xi_{cls^{syn}_{1}}$ are the coefficients for the input images and the generated images, respectively. 
When Phase 1 is finished, the initial multivariate GM distribution of the latent features can be obtained and denoted by $M_{init}$ and $\Sigma_{init}$.

\subsection{Phase 2: Local Enhancement for Image Synthesis}
\begin{figure}[htbp]
  \centering 
  \includegraphics[width=0.95\linewidth]{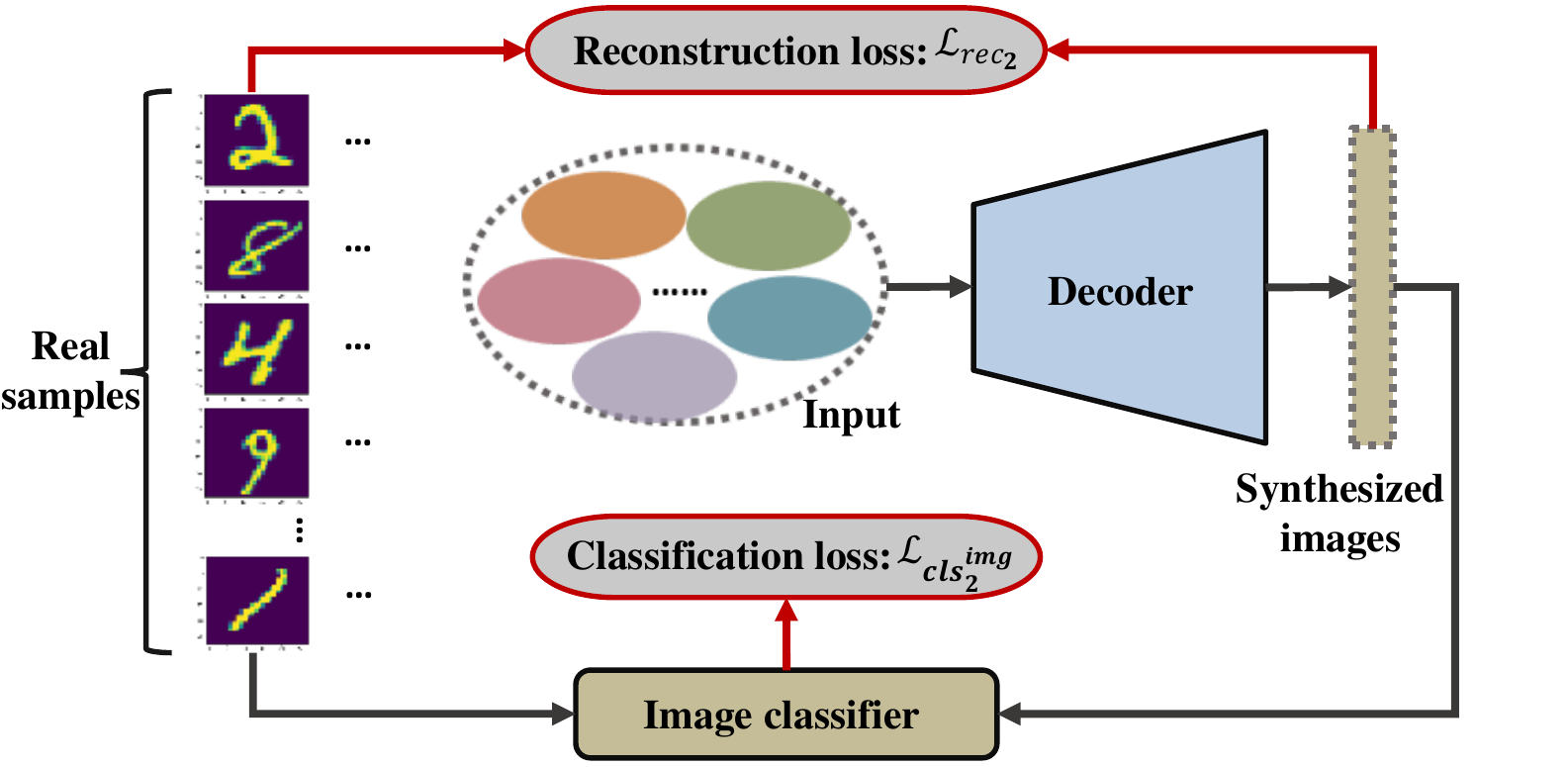}\\
  \caption{Phase 2: the decoder and the image classifier are locally trained.}
  \vspace{-0.2cm}
\end{figure}
In Phase 2, we focus on enhancing the ability of the decoder and the image classifier. 
As shown in Fig. 3, the latent features are firstly produced as the input of the decoder by sampling from the learned distribution with the mean $M_{init}$ and the covariance $\Sigma_{init}$. 
Based on the synthesized images and the real samples randomly chosen according to the number and labels of the input, the reconstruction loss $\mathcal{L}_{rec_{2}}$ and classification $\mathcal{L}_{cls^{img}_{2}}$ are calculated.
Hence, the total loss $\mathcal{L}_{ph_{2}}$ of this phase is defined as:
\begin{equation}
\begin{aligned}
\mathcal{L}_{ph_{2}}=&\beta_{rec_{2}}*\mathcal{L}_{rec_{2}}+
\beta_{cls^{img}_{2}}*\mathcal{L}_{cls^{img}_{2}}
\end{aligned}
\end{equation}
\begin{equation}
\begin{aligned}
\mathcal{L}_{rec_{2}}=\mathcal{L}_{rec_{2}\_min}+\xi_{rec_{2}}*\mathcal{L}_{rec_{2}\_maj}
\end{aligned}
\end{equation}  
\begin{equation}
\begin{aligned}
\mathcal{L}_{cls^{img}_{2}}=&\mathcal{L}_{cls^{ori}_{2}\_min}+\xi_{cls^{ori}_{2}}*\mathcal{L}_{cls^{ori}_{2}\_maj}\\
&\mathcal{L}_{cls^{syn}_{2}\_min}+\xi_{cls^{syn}_{2}}*\mathcal{L}_{cls^{syn}_{2}\_maj}
\end{aligned}
\end{equation} 
where $\beta_{rec_{2}}$ and $\beta_{cls^{img}_{2}}$ are the corresponding weighting coefficients.
$\xi_{rec_{2}}$ is to adjust the weight between the majority and the minority samples for reconstruction loss.
Both the real samples and the generated images are sent to the image classifier for calculating the classification loss, with $\xi_{cls^{ori}_{2}}$ and $\xi_{cls^{syn}_{2}}$ adjusting their weights between the majority and the minority.

\subsection{Phase 3: Local Enhancement for Feature Learning}
\begin{figure}[htbp]
  \centering 
  \includegraphics[width=0.95\linewidth]{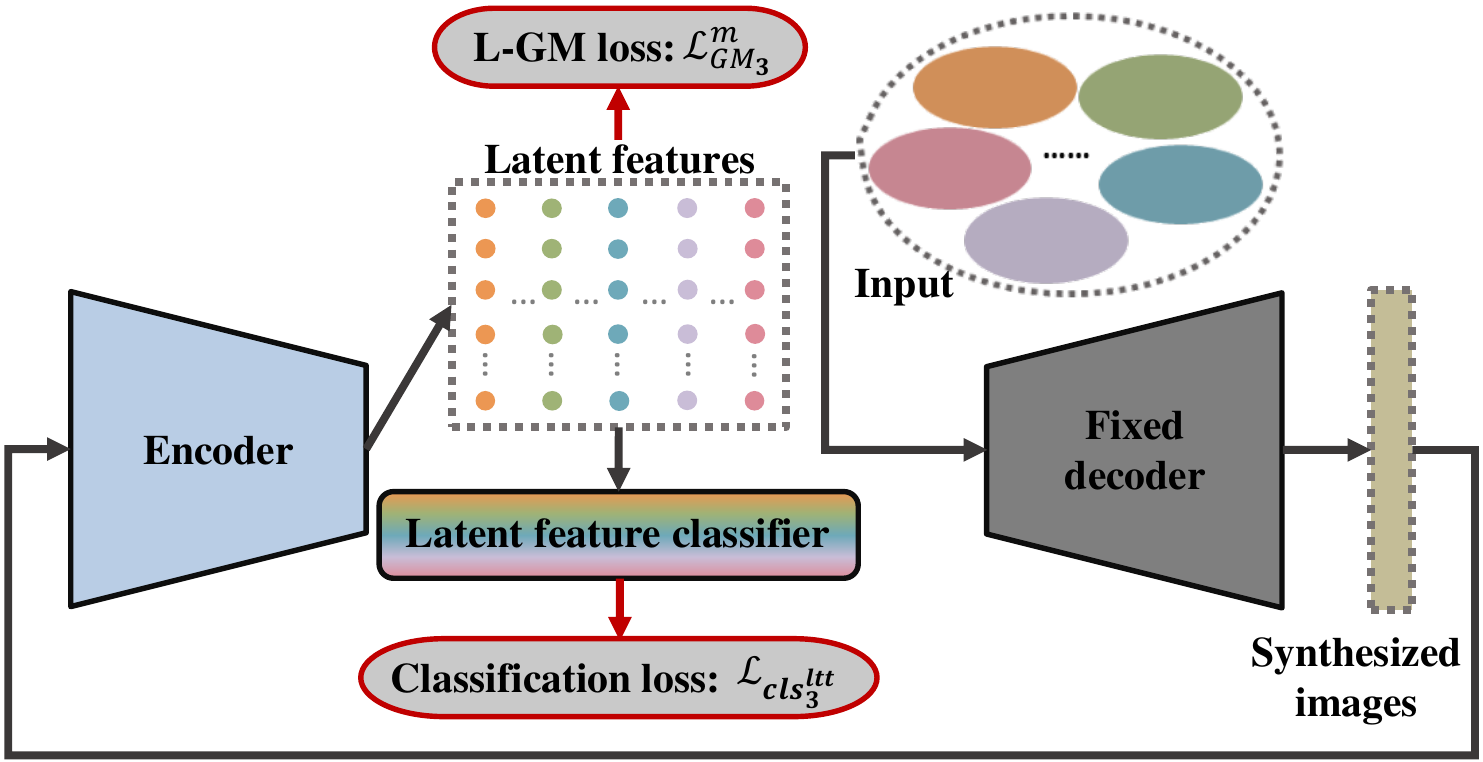}\\
  \caption{Phase 3: the encoder and the latent feature classifier are locally trained in this phase, where the decoder parameters are kept fixed.}
  \vspace{-0.2cm}
\end{figure}

The specific operation of Phase 3 is depicted in Fig. 4.
This phase occurs two kinds of latent features: 
one is sampled from the current GMM learned from Phase 1 and treated as the input of the whole module; another is encoded from the synthesized images, which are decoded from the input one. 
To further optimize the parameters of the encoder and the latent feature classifier, both the L-GM loss and the classification loss are calculated based on the latter latent features.
The total loss is defined by:
\begin{equation}
\begin{aligned}
\mathcal{L}_{ph_{3}}=\beta_{GM_{3}}*\mathcal{L}_{GM_{3}}^{m}
+\beta_{cls^{ltt}_{3}}*\mathcal{L}_{cls^{ltt}_{3}}
\end{aligned}
\end{equation} 
where $\mathcal{L}_{GM_{3}}^{m}$ and $\mathcal{L}_{cls^{ltt}_{3}}$ are the L-GM loss and the latent feature classification loss in this phase, $\beta_{GM_{3}}$ and $\beta_{cls^{ltt}_{3}}$ are the corresponding weight coefficients.
Since the input data's distribution learned from Phase 1 already considers the bias between the majority and the minority, we ignore the influence of skewed distribution from the imbalanced data for each loss in this phase. 

\subsection{Phase 4: Evolution of Latent Feature Distribution}
\begin{figure*}[h]
  \centering 
  \includegraphics[width=0.95\linewidth]{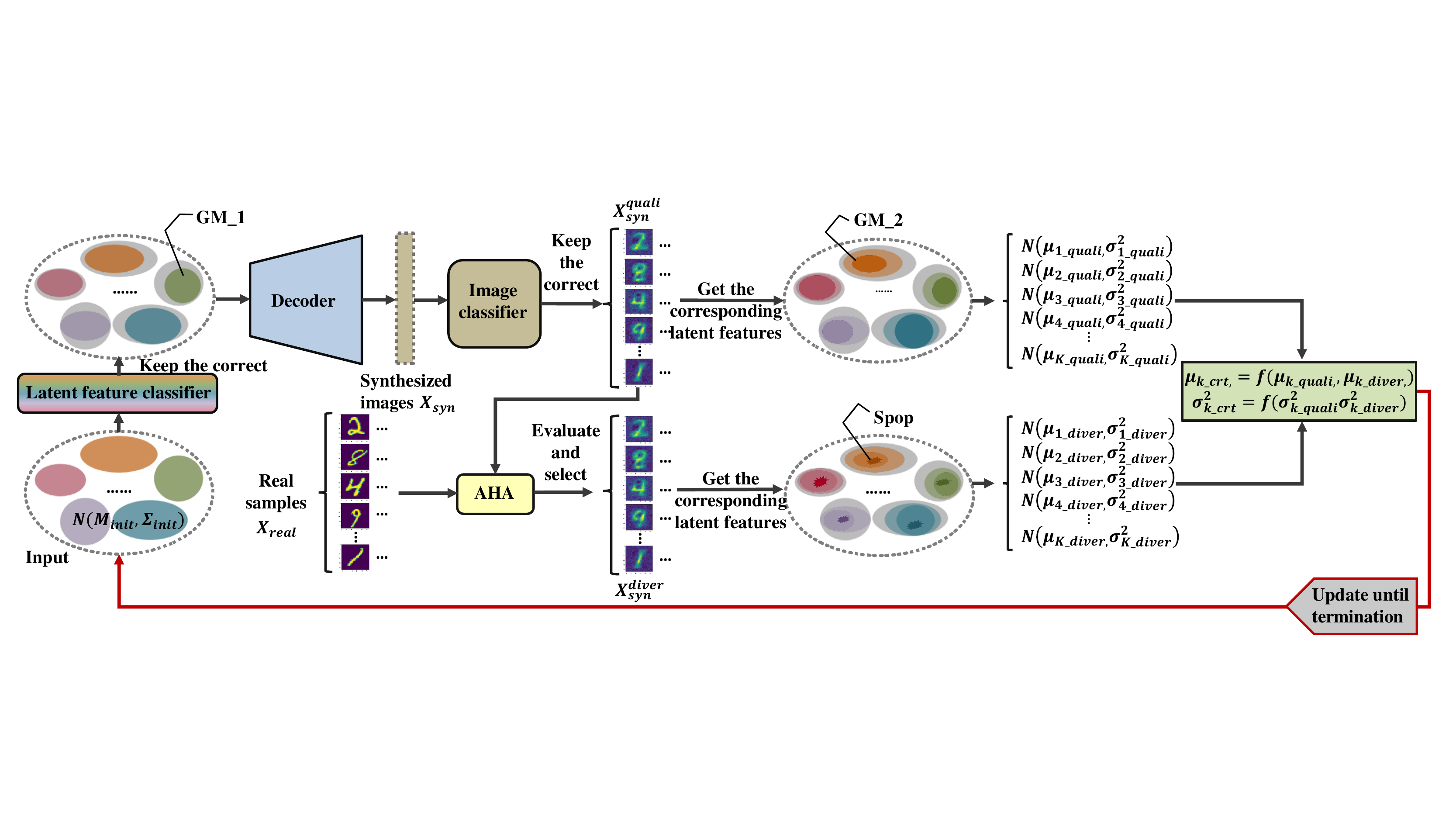}\\
  \caption{Phase 4: the MEDA evolves the GM distributions of the latent features, where the function $f$ denotes the evolution rule.}
  \vspace{-0.2cm}
\end{figure*}

The multivariate GM distribution of the latent features is evolved for better quality and diversity in Phase 4, and the specific details are depicted in Fig. 5.
The input is formed by randomly sampling from the current GMM $\mathcal{N}(M_{init}, \Sigma_{init})$ and is firstly passed through the latent feature classifier.
Then the correctly classified ones are kept and denoted by `GM\_1', where the coloured part indicates the correct ones while the wrong part turns grey.
Next, the correct is decoded to the reconstructed images, denoted as $X_{syn}$, and the images classified correctly are reserved and named $X_{syn}^{quali}$.
To this point, $X_{syn}^{quali}$ is considered to be high-quality images, meaning that they can be easily identified by the human eye or recognized by an image classifier. 

However, the diversity of the generated images can not be ensured, which may cause an overfitting problem for classifiers.
A fitness function is designed based on the similarities among samples to increase the diversity by selecting the diverse generations.
The similarities are calculated by the average hash algorithm (AHA) \cite{weng2011secure} between $X_{syn}^{quali}$ and the real samples $X_{real}$ randomly selected from the imbalanced data: (1) compute the hash values for all the pairs of samples, with each pair consisting of one synthesized image and one real image; (2) compare the hash values to get the similarities; (3) Sort the similarities and select the corresponding synthesized samples with the lowest similarities based on selection rate as superior individuals, named $X_{syn}^{diver}$.

To guide the evolution direction to better diversity based on the high quality, the corresponding latent features of $X_{syn}^{quali}$ and $X_{syn}^{diver}$ are found and noted by `$GM\_2$' and `$Spop$' as shown in Fig. 5.
Compared with the traditional EDAs, the distribution of the current latent features is replaced with that of `$GM\_2$' when evolving to guarantee the quality of synthesized samples. Simultaneously, 
`$Spop$' contributes to optimizing the latent feature distribution to be more diverse. 
Therefore, the MEDA updates the GM distribution by:
\begin{equation}
\begin{aligned}
\mu_{k\_crt}=\gamma\mu_{k\_quali} + (1-\gamma)\mu_{k\_diver} 
\end{aligned}
\end{equation} 
\begin{equation}
\begin{aligned}
\sigma_{k\_crt}^{2}=\gamma\sigma^{2}_{k\_quali} + (1-\gamma)\sigma^{2}_{k\_diver}
\end{aligned}
\end{equation} 
where $\mu_{k\_crt}$ and $\sigma_{k\_crt}^{2}$ calculate the mean and the covariance of the latest Gaussian distribution for the latent features in class $k$.
$\mu_{k\_quali}$ and $\sigma_{k\_quali}^{2}$ denote the mean and the covariance of class $k$ in `$GM\_2$', respectively, while $\mu_{k\_diver}$ and $\sigma_{k\_diver}^{2}$ are those of class $k$ in `$Spop$'.
$\gamma$ represents the weighting coefficients.
 
Each update is done, the latent features are re-sampled from the current GM distribution, and the steps above are repeated until the termination condition is met.

\subsection{Program of Four Phases}
\begin{algorithm*}[h]
\caption{MEDA\_LUDE training procedure}
\label{alg1}
\begin{multicols}{2}
\begin{algorithmic}[1]
  \renewcommand{\algorithmicrequire}{       \textbf{Input:}}
  \REQUIRE  $X=\lbrace X_{maj},X_{min} \rbrace$,$Y=\lbrace Y_{maj},Y_{min} \rbrace$
  \renewcommand{\algorithmicensure}{       \textbf{Output:}}
  \ENSURE $\theta_{de}$, $M_{opti}$, $\Sigma_{opti}$
  \\ \hspace*{\fill} \\
  \STATE /*Phase 1*/
  \WHILE{not converged}
  \FOR{each mini-batch data}
  \STATE update $\theta_{en}$ with $\mathcal{L}_{rec_{1}}$, $\mathcal{L}_{GM_{1}}^{m}$, $\mathcal{L}_{cls^{ltt}_{1}}$, and $\mathcal{L}_{cls^{img}_{1}}$
  \STATE update $\theta_{de}$ with $\mathcal{L}_{rec_{1}}$, $\mathcal{L}_{cls^{img}_{1}}$
  \STATE update $\theta_{cls\_ltt}$ with $\mathcal{L}_{cls^{ltt}_{1}}$
  \STATE update $\theta_{cls\_img}$ with $\mathcal{L}_{cls^{img}_{1}}$
  \STATE update $M$, $\Sigma$ with $\mathcal{L}_{GM_{1}}^{m}$
  \ENDFOR
  \ENDWHILE
  \STATE $M_{init}=M$, $\Sigma_{init}=\Sigma$
  \\ \hspace*{\fill} \\
  \FOR{number of training iterations}
  \STATE /*Phase 2*/
  \WHILE{not converged}
  \FOR{each mini-batch data}
  \STATE update $\theta_{de}$ with $\mathcal{L}_{rec_{2}}$, $\mathcal{L}_{cls^{img}_{2}}$
  \STATE update $\theta_{cls\_img}$ with $\mathcal{L}_{cls^{img}_{2}}$
  \ENDFOR
  \ENDWHILE
  \\ \hspace*{\fill} \\
  \STATE /*Phase 3*/
  \WHILE{not converged}
  \FOR{each mini-batch data}
  \STATE update $\theta_{en}$ with $\mathcal{L}_{GM_{3}}^{m}$, $\mathcal{L}_{cls^{ltt}_{3}}$
  \STATE update $\theta_{cls\_ltt}$ with $\mathcal{L}_{cls^{ltt}_{3}}$
  \ENDFOR
  \ENDWHILE
  \\ \hspace*{\fill} \\
  \STATE /*Phase 4*/
  \STATE $FEAT_{4}=Sample(\mathcal{N}(M_{init},\Sigma_{init}))$ 
  \WHILE{halting condition is not met}
  \STATE $\widehat{label}_{ltt}=Classify\_Ltt(FEAT_{4}, \theta_{cls\_ltt})$
  \STATE $GM\_{1}=Select\_Correct(FEAT_{4}, \widehat{label}_{ltt}, label_{ltt})$
  \STATE $X_{syn}=Decode(GM\_{1}, \theta_{de})$
  \STATE $\widehat{label}_{img}=Classify\_Img(X_{syn}, \theta_{cls\_img})$
  \STATE $X_{syn}^{quali}=Select\_Correct(X_{syn}, \widehat{label}_{img}, label_{img})$ 
  \STATE $fitness=Cal\_Fitness(X_{syn}^{quali}, X_{real}, AHA)$
  \STATE $X_{syn}^{diver}=Evaluate\_Select(X_{syn}^{quali}, fitness)$
  \STATE $GM\_2=Find\_Ltt(X_{syn}^{quali})$\\
  $Spop=Find\_Ltt(X_{syn}^{diver})$ 
  \STATE $\mathcal{N}(\mu_{k\_quali},\sigma^{2}_{k\_qauli})=Cal\_Gaussian(GM\_2)$\\
  $\mathcal{N}(\mu_{k\_diver},\sigma^{2}_{k\_diver})=Cal\_Gaussian(Spop)$
  \STATE $\mathcal{N}(\mu_{k\_crt},\sigma^{2}_{k\_crt})=Evolve(\mu_{k\_quali}, \mu_{k\_diver},$\\ $\sigma^{2}_{k\_qauli},\sigma^{2}_{k\_diver})$
  \STATE $M_{crt}=[\mu_{1\_crt};\mu_{2\_crt}\ldots;\mu_{k\_crt}]$\\
  $\Sigma_{crt}=[\sigma^{2}_{1\_crt};\sigma^{2}_{2\_crt}\ldots;\sigma^{2}_{k\_crt}]$
  \STATE $FEAT_{4}=Sample(\mathcal{N}(M_{crt},\Sigma_{crt}))$
  \ENDWHILE
  \STATE $M_{opti}=M_{crt}$
  \STATE $\Sigma_{opti}=\Sigma_{crt}$
  \ENDFOR
  \\ \hspace*{\fill} \\
  \RETURN $\theta_{de}$, $M_{opti}$, $\Sigma_{opti}$
\end{algorithmic}
\end{multicols}
\end{algorithm*}

Four phases mentioned above are programmed to train the MEDA\_LUDE algorithm, and the pseudocode is shown in Algorithm 1.

In Algorithm 1, the input of the MEDA\_LUDE is composed of the majority samples $X_{maj}$ and the minority samples $X_{min}$, namely $X=\lbrace X_{maj},X_{min} \rbrace$. Their corresponding labels are denoted as $Y=\lbrace Y_{maj},Y_{min} \rbrace$.
The model parameters of the encoder, the decoder, the latent feature classifier, and the image classifier are signified as $\theta_{en}$, $\theta_{de}$, $\theta_{cls\_ltt}$, and $\theta_{cls\_img}$, respectively.

Besides, Algorithm 1 makes use of the follow functions in Phase 4:
\begin{itemize}
\item $Sample(distribution)$: sample from `$distribution$' to generate population.
\item $Classify\_Ltt(ltt, para)$: classify latent features `$ltt$' by latent feature classifier with model parameters `$para$'.
\item $Select\_Correct(data, label1, label2)$: select the correctly classified `$data$' according to the classification results `$label1$' and the real labels `$label2$'.
\item $Decode(latent feature, para)$: decode `$latent feature$' to synthesized images by decoder with model parameters `$para$'.
\item $Classify\_Img(img, para)$: classify `$img$' by image classifier with model parameters `$para$'.
\item $Cal\_Fitness(data1, data2, algorithm)$: calculate fitness of `$data1$' based on `$data2$' by `$algorithm$'.
\item $Evaluate\_Select(data, fitness)$: evaluate `$data$' based on `$fitness$' and select promising solutions.
\item $Find\_Ltt(data)$: find corresponding latent features of `$data$'.
\item $Cal\_Gaussian(data)$: give Gaussian distribution of `$data$'.
\item $Evolve(mean1, mean2, cov1, cov2)$: evolve Gaussian distribution based on `$mean1$', `$mean2$', `$cov1$', `$cov2$' to the update distribution $\mathcal{N}(mean_{crt}, cov_{crt})$ by:
\begin{equation}
\begin{aligned}
mean_{crt}=\gamma mean1 + (1-\gamma) mean2
\end{aligned}
\end{equation} 
\begin{equation}
\begin{aligned}
cov_{crt}=\gamma cov1 + (1-\gamma) cov2
\end{aligned}
\end{equation} 
where $\gamma$ is the coefficient in Eq. (25) and Eq. (26).
\end{itemize}
In Phase 4, $label_{ltt}$ and $label_{img}$ denote the real labels of $FEAT_{4}$ and $X_{syn}$, respectively.

\section{Experiments}
This section evaluates the proposed MEDA\_LUDE algorithm on the imbalanced datasets randomly formed from MNIST and CIFAR-10 datasets. Furthermore, the MEDA\_LUDE is also applied to the industrial field and has achieved significant success. The experiments are implemented with NVIDIA RTX 2080Ti, 64 GB RAM, Intel(R) Xeon(R) Silver 4114 CPU, Ubuntu 18.04.5 LTS, Pytorch 1.7.0.

\subsection{Evaluation on Benchmark-based Imbalanced Datasets}
Two random seeds are set for each benchmark dataset to allocate the categories of the majority samples and the minority samples.

For the MNIST dataset, the handwritten digits from 0 to 9 are first symbolized with numbers from 0 to 9. Then we randomly select five numbers as the minority classes
with the random seed function in the NumPy library for Python programming language. The seeds are set to be 0 and 5, respectively, where the minority classes' numbers are 2, 8, 4, 9, 1 when the seed is 0, and are 9, 5, 2, 4, 7 when the seed is 5 accordingly. The two imbalanced datasets built from the MNIST dataset are both split into the training and validation sets. The training set contains 50 samples for each minority class and 5000 examples for each majority class. The validation set includes 4000 samples uniformly distributed within ten categories, and the test set is the same as that of the original MNIST dataset. Hence, the MEDA\_LUDE is finally experimented on two randomly formed datasets with an imbalanced ratio \cite{kang2016noise} being 100.

For the CIFAR-10 dataset, the categories of the airplane, automobile, bird, cat, deer, dog, frog, horse, ship, and truck are marked with numbers 0, 1, 2, 3, 4, 5, 6, 7, 8, and 9, accordingly. Here, the random seeds are set to be 1 and 7, corresponding to the minority classes being bird, truck, frog, deer, airplane, and ship, dog, airplane, bird, and automobile, respectively.
From the original training set of the CIFAR-10 dataset, 45 and 4500 examples are selected for each minority class and each majority class, respectively, constituting imbalanced datasets with an imbalanced ratio of 100 under two different circumstances.
Among the remaining samples of the training set, 400 samples are chosen randomly for each class to form a validation set. The test set is the original one of the CIFAR-10 dataset.
 \begin{table*}[!h]\footnotesize
\renewcommand\arraystretch{1}
\centering
\caption{Classification performance of the comparative methods for MNIST}
\begin{tabular}{p{1.2cm} p{0.6cm}  p{1cm}p{1cm} p{1cm} p{1cm} p{1cm} p{1.2cm} p{1.2cm}}
  \hline
  \hline
   Evaluation & Seed & RMR& ROS  & SMOTE & ADASYN  & CVAE & CVAE\_& \textbf{MEDA\_}\\ 
   Critieria&  &   & &  &   &  & SeTred&\textbf{LUDE}\\
  \Xhline{0.8pt}
   Accuracy & 0&93.89&94.63&94.67&95.31&95.37&95.58&\textbf{95.88}\\
  (\%)&5&92.28&93.08&94.19&93.48&95.07&95.44&\textbf{95.49}\\
  \hline
   Precision&0&94.20&94.95&94.94&95.45&95.57&95.81&\textbf{96.05}\\
    (\%)&5&93.07&93.81&94.33&93.86&95.35&95.62&\textbf{95.67}\\
  \hline
   Recall &0&93.89&94.63&94.67&95.31&95.37&95.58&\textbf{95.88}\\
    (\%) &5&92.28&93.08&94.19&93.48&95.07&95.44&\textbf{95.49}\\
  \hline
      Specificity &0&99.30&99.41&99.41&99.48&99.48&99.51&\textbf{99.55}\\   
     (\%)&5&99.14&99.24&99.36&99.27&99.45&\textbf{99.49}&\textbf{99.49}\\
  \hline
   F1&0&93.81&94.58&94.60&95.28&95.33&95.56&\textbf{95.86}\\
    (\%)&5&92.21&93.05&94.14&93.42&95.05&95.42&\textbf{95.46}\\
  \hline
   GM &0&96.49&96.94&96.96&97.35&97.37&97.50&\textbf{97.67}\\ 
    (\%)&5&95.54&96.02&96.70&96.27&97.20&97.42&\textbf{97.44}\\
  \hline
  AUC &0&0.9932&0.9954&0.9958&0.9936&0.9962&0.9961&\textbf{0.9966}\\
        &5&0.9937&0.9946&0.9941&0.9946&0.9953&0.9953&\textbf{0.9961}\\
  \hline
  \hline
\end{tabular}
\end{table*}

\begin{table*}[!h]\footnotesize
\renewcommand\arraystretch{1}
\centering
\caption{Classification performance of the comparative methods for Cifar10}
\begin{tabular}{p{1.2cm} p{0.6cm}  p{1cm}p{1cm} p{1cm} p{1cm} p{1cm} p{1.2cm} p{1.2cm}}
  \hline
  \hline
   Evaluation & Seed & RMR& ROS  & SMOTE & ADASYN  & CVAE & CVAE\_& \textbf{MEDA\_}\\ 
   Critieria&  &   & &  &   &  & SeTred&\textbf{LUDE}\\
  \Xhline{0.8pt}
   Accuracy &1&48.78&49.15&50.35&50.94&51.06&51.36&\textbf{52.65}\\
  (\%)&7&48.61&49.18&49.43&50.26&51.10&52.22&\textbf{52.96}\\
  \hline
   Precision&1&60.49&63.02&59.61&62.58&61.35&61.10&\textbf{63.42}\\
    (\%)&7&60.52&66.17&64.54&64.34&61.27&64.17&\textbf{66.85}\\
  \hline
   Recall&1&48.78&49.15&50.35&50.94&51.06&61.36&\textbf{52.65}\\
    (\%)&7&48.61&49.18&49.43&50.26&51.10&52.22&\textbf{52.96}\\
  \hline
      Specificity &1&94.31&94.35&94.48&94.55&94.56&94.60&\textbf{94.74}\\   
     (\%)&7&94.29&94.35&94.38&94.47&94.57&94.69&\textbf{94.77}\\
  \hline
   F1&1&41.12&41.72&45.87&44.97&43.12&46.00&\textbf{47.04}\\
    (\%)&7&45.52&42.77&43.01&40.94&45.89&45.46&\textbf{47.35}\\
  \hline
   GM &1&60.02&60.22&64.02&63.84&61.73&64.67&\textbf{64.70}\\ 
    (\%)&7&62.67&60.01&60.92&57.90&63.86&62.27&\textbf{63.94}\\
  \hline
  AUC &1&0.8121&0.7985&0.7972&0.8295&0.7990&0.8212&\textbf{0.8406}\\
        &7&0.7889&0.7814&0.7877&0.7914&0.8052&\textbf{0.8129}&0.8094\\
  \hline
  \hline
\end{tabular}
\end{table*}

To validate the effectiveness of our proposed method, 
all the methods first synthesize samples to construct the balanced dataset based on the imbalanced training set.
Then the final classifier is trained on the balanced training set and tested to achieve the final classification performance.
The closer the constructed balanced training data's distribution is to the real data's, the better the final classification performance will be. The classification performance on imbalanced MNIST and CIFAR-10 datasets are shown in TABLE \uppercase\expandafter{\romannumeral1} and TABLE \uppercase\expandafter{\romannumeral2}, respectively.
In TABLE \uppercase\expandafter{\romannumeral1}, the MEDA\_LUDE algorithm obtains the optimal results on various indexes under two random seeds among all the comparative methods, indicating that the samples generated by our method are superior to its competitors in quality-diversity trade-off and hence contributes to training the classifier.
On the CIFAR-10 dataset, the proposed MEDA\_LUDE method shows the competitive classification performance, except that the best value for criterion `AUC' is obtained by the CVAE\_SeTred method when the random seed is 7.
To conclude, the optimization phases we design for the latent feature distribution and the whole neural networks are efficacious to trade off the quality and diversity of synthesized samples.

The generated minority samples for the MNIST dataset are visualized with t-SNE \cite{van2008visualizing}, as depicted in Fig. 6. 
The `Original' gives the distribution of the real minority data, and the others represent the distribution of the minority samples synthesized by their corresponding method, where the `MEDA\_LUDE' exhibits the generated data's distribution when the random seed is 5.
Since ADASYN, SMOTE, and ROS provide a balanced training set directly, the generated samples need to be chosen randomly from the minorities of the balanced set with a certain amount.
As Fig. 6 shows, the distribution of samples synthesized by `ROS' is a disorder without distinct boundaries, and the distribution corresponding to `RMR' is far from the original distribution.
In contrast, the distributions from `ADASYN' and `SMOTE' are relatively reasonable but appear locally aggregated and integrally sparse.
Compared with `ADASYN' and `SMOTE', the minority samples generated from three generative models, namely `CVAE', `CVAE\_SeTred', and `MEDA\_LUDE', are evenly distributed for each class, which accord with the original samples better.
Obviously, the distribution learned from `CVAE' is the most limited, and the second is that of `CVAE\_SeTred'. With the same amount of the minority examples,  our proposed `MEDA\_LUDE' has learned more broad distribution, validating the better diversity of our synthesized samples indirectly.

\begin{figure}[htbp]
  \centering 
  \includegraphics[width=0.9\linewidth]{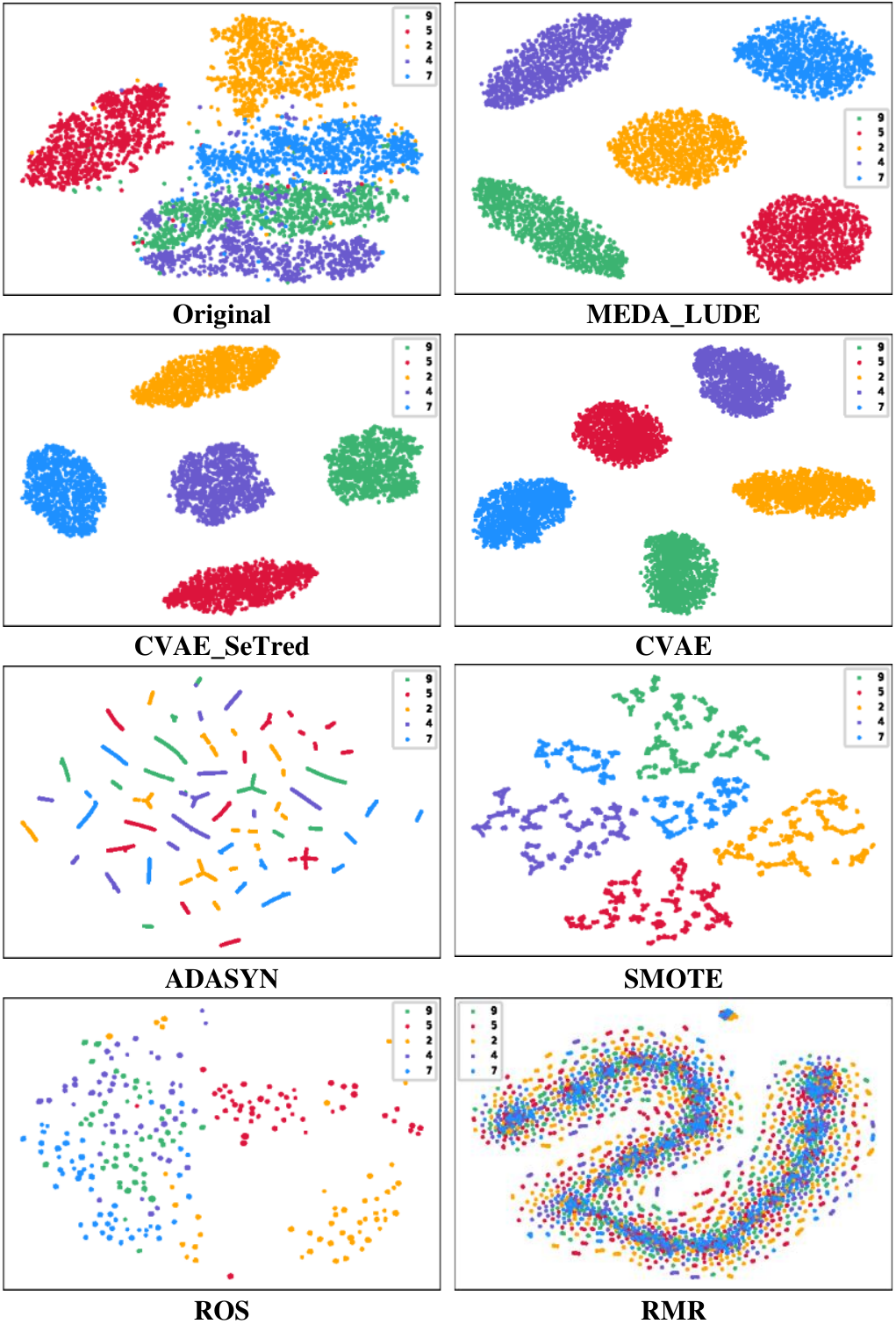}\\
  \caption{t-SNE visualizations of the generated minority samples for MNIST dataset}\label{Fig. 1}
  \vspace{-0.2cm}
\end{figure}

\begin{figure}[htbp]
  \centering 
  \includegraphics[width=0.9\linewidth]{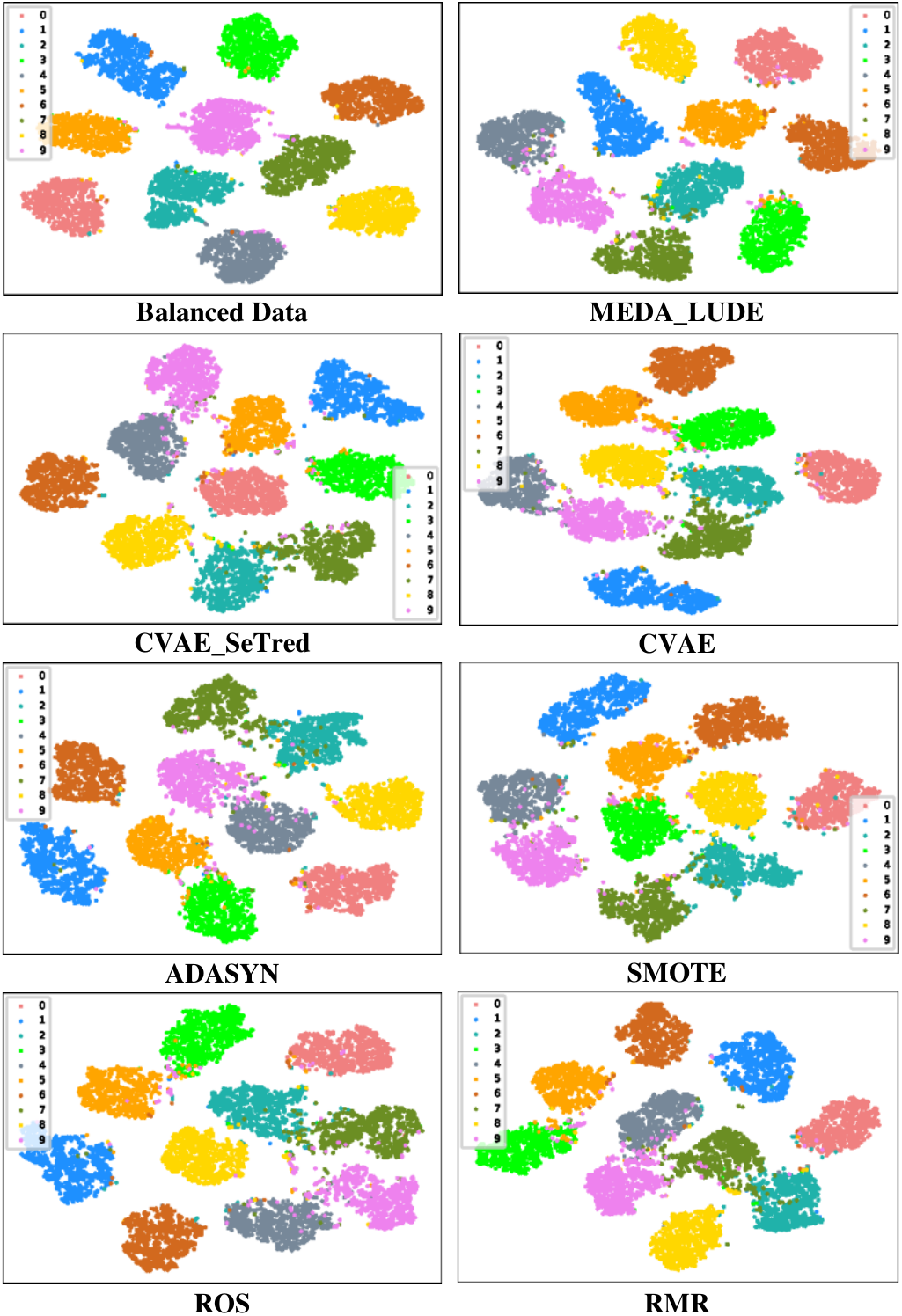}\\
  \caption{t-SNE visualizations of the features for MNIST dataset}\label{Fig. 1}
  \vspace{-0.2cm}
\end{figure} 

In addition to the synthesized minority samples, we also visualize the features output from the last fully connected layer of the final classifier, which is trained on the balanced data formed by various methods. It should be noted that the visualized features correspond to the inputs of the test set instead of the training data.
The visualizations are depicted in Fig. 7, where the `Balanced data' describes the feature distributions mapped from the test data by the classifiers trained on the original balanced training data. Different from the distributions visualized from the generated samples, the closer the points of the same class are, and the clearer the clusters' boundaries of the different classes appear, the higher probability the samples are classified correctly with.
Compared with the `Balanced data', the clusters of the different classes visualized from the other methods have obvious overlaps, which denotes the place where samples are classified wrongly.
Among all of these synthetic methods, the overlaps of the `MEDA\_LUDE' are relatively less, and the boundaries of different categories are more apparent.
In view of this, our proposed MEDA\_LUDE algorithm can generate samples whose distributions are closer to the real samples than the comparative methods.
As a result, the classification performance of the classifier trained on the balanced data the MEDA\_LUDE  constructs improves a lot.

\subsection{Industrial Application with Real-world Datasets}
The MEDA\_LUDE algorithm is also applied to the industrial field successfully. In the textile production process, fabric defect classification is critical to control the textile quality. However, during the real-world data acquisition, imbalanced problems are common due to the factor of production equipment and the characteristics of fabric defects, which sharply weaken the classification performance and significantly influence the fabric quality.
To solve the imbalanced classification of fabric defects in practical industrial applications, we firstly collect and process fabric defect data and then construct fabric defect datasets: DHU-FD and ALIYUN-FD. 
\begin{figure}[htbp]
  \centering 
  \includegraphics[width=1\linewidth]{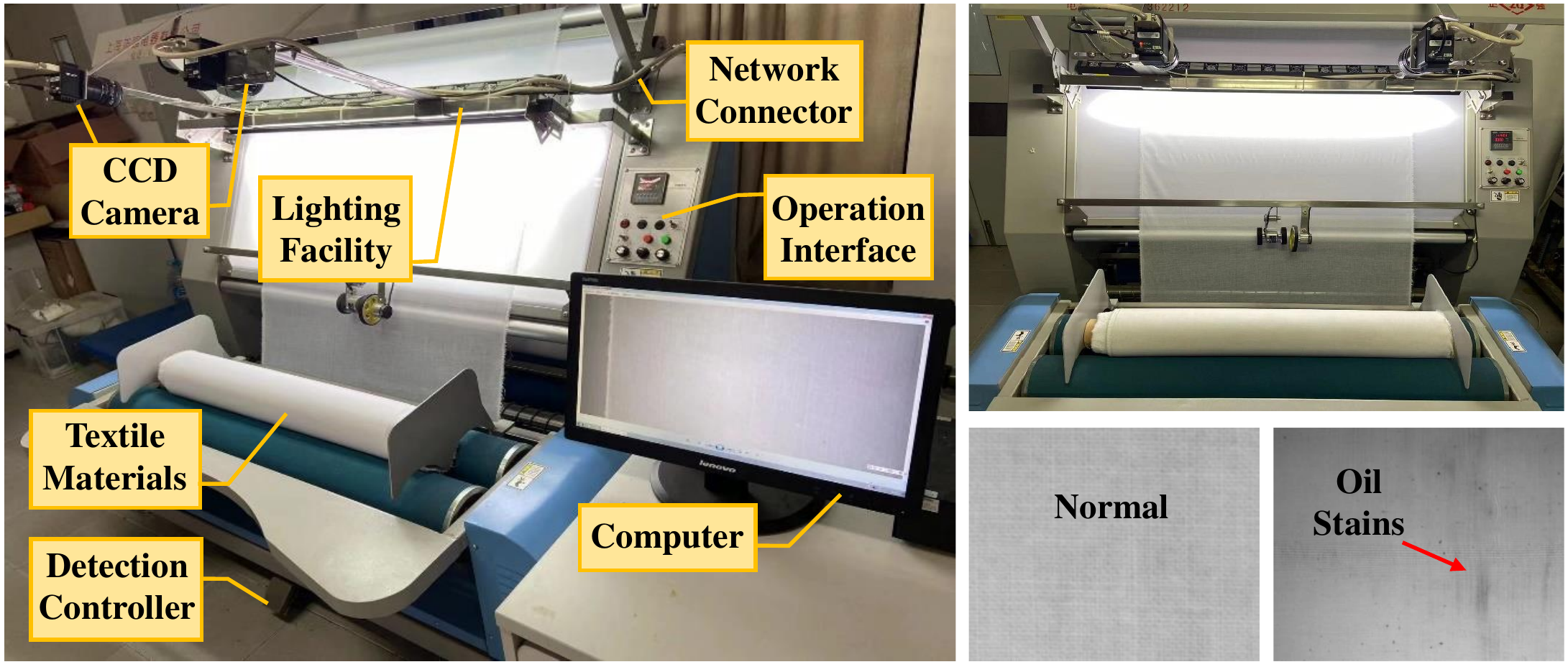}\\
  \caption{Hardware acquisition device and raw textile image examples}\label{Fig. 1}
  \vspace{-0.2cm}
\end{figure}

\begin{figure}[htbp]
  \centering 
  \includegraphics[width=0.94\linewidth]{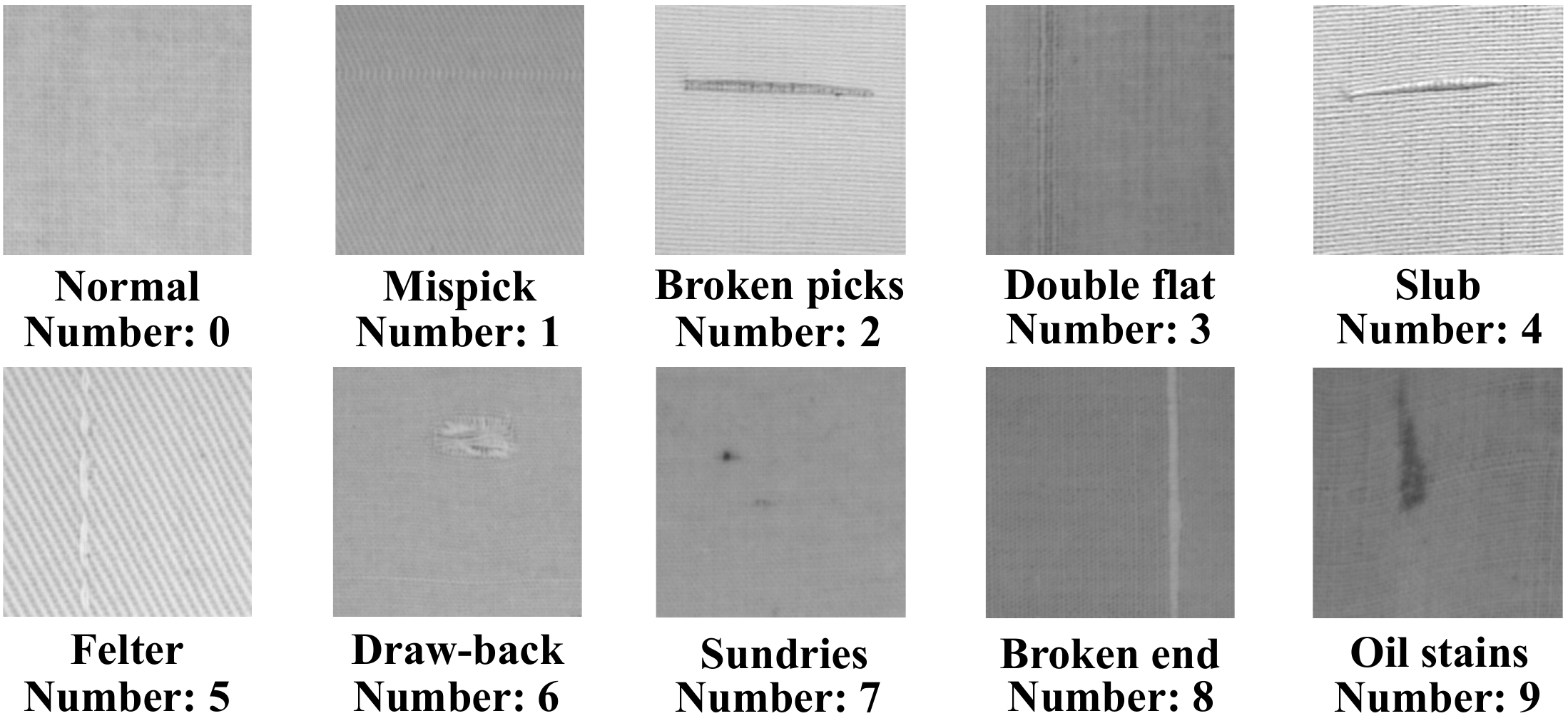}\\
  \caption{Examples and numbers for DHU-FD}\label{Fig. 1}
  \vspace{-0.2cm}
\end{figure}

\begin{figure}[htbp]
  \centering 
  \includegraphics[width=.95\linewidth]{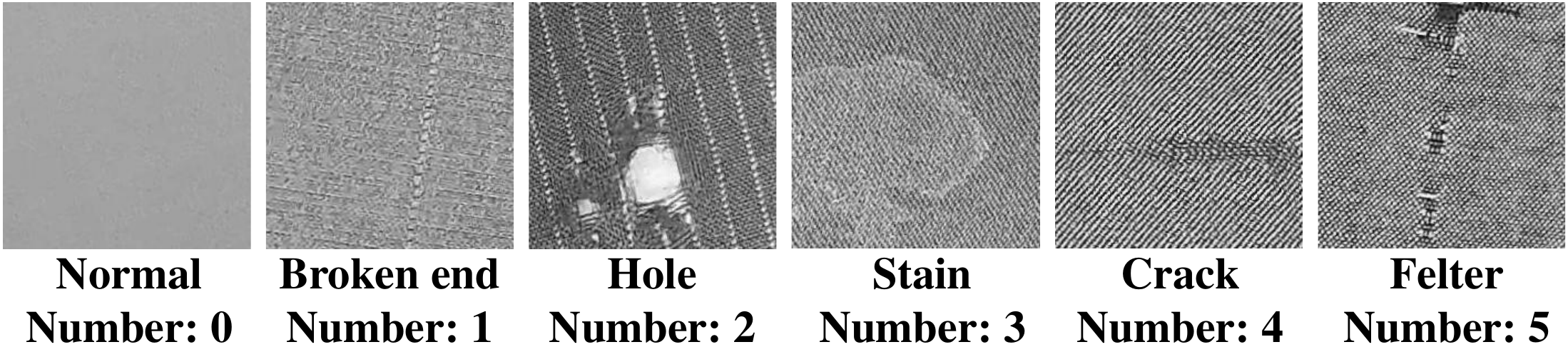}\\
  \caption{Examples and numbers for ALIYUN-FD}\label{Fig. 1}
  \vspace{-0.2cm}
\end{figure}

For DHU-FD, the fabric defect data is collected from Donghua University (DHU) based on the hardware acquisition device, including two CCD cameras, a network connector, lighting facility, computer, operation interface, detection controller, and textile materials. The acquisition device is shown in Fig. 8, where the right-bottom corner gives examples of raw fabric textiles labelled as normal and oil stains. The original DHU textile images are in $1280*1024$ pixels.
Then the acquired data is processed with Adobe Photoshop CS6 by intercepting the defected part of unified pixels, and each intercepted image only contains one type of defect. 
These images are labelled according to the characteristic descriptions of fabric defects. 
Finally, the DHU-FD dataset is constructed with ten classes, and each class contains 100 samples of size $220*220$.
Types in the sequence of `Normal', `Mispick', `Broken picks', `Double flat', `Slub', `Felter', `Draw-back', `Sundries', `Broken end', and `Oil stains' are numbered  from 0 to 9, respectively. Examples, labels, and numbers for ten types of fabric defects are given in Fig. 9.

In addition to the DHU data, we also download fabric image data that is public online in the Tianchi competition sponsored by Alibaba. The size of the original fabric data is $2560*1920$ pixels. After the same operations in processing raw images, the ALIYUN-FD dataset is constructed, including six types, with each keeping 1500 samples. 
Images in the ALIYUN-FD dataset are in $220*220$ pixels. 
Labels in the order of Normal', `Broken end', `Hole', `Stain', `Crack', and `Felter' are sequentially numbered from 0 to 5.
The specific details of ALIYUN-FD examples are shown in Fig. 10.

For both the DHU-FD dataset and the ALIYUN-FD dataset, two random seeds are set to determine minority classes.
For the DHU-FD dataset, the random seeds are set to 1 and 3 to select five classes as the minority categories from all ten types.
The corresponding minority types are `Broken picks', `Oil stains', `Draw-back', `Slub', `Normal' when the seed is 1, and being `Felter', `Slub', `Mispick', `Broken picks', `Oil stains' when the seed is 3.
In the randomly formed imbalanced DHU-FD dataset, each minority class contains 6 samples, while each majority class has 60 ones in the training set. The validation and test sets include 200 images evenly distributed among ten classes.
For the ALIYUN-FD dataset, the random seeds are 0 and 1. Correspondingly, seed 0 selects `Felter', `Hole', and `Broken end' as the minority classes, while seed 1 chooses `Hole', `Broken end', and `Crack' as the minority types. The training set comprises of 270 minority samples and 2700 majority samples, where the imbalance ratio is 10. In the validation set, 300 images are included in each class, and the same goes for the test set.

To observe the situation of training the MEDA\_LUDE algorithm in large-size images, the training loss curves varying iterations are plotted for Phase 1, Phase 2, and Phase 3, since Phases 1-3 are responsible for training models, while Phase 4 aims at evolving latent features. 
The Adam optimizer is adopted, and the activation function is ReLU.
In Phase 1, the whole model is firstly trained, including the encoder, the decoder, the latent feature classifier, and the image classifier. Hence the total loss consists of reconstruction loss, L-GM loss, latent feature classification loss, and image classification loss, as depicted in Fig.11.  It can be seen that all the losses in Phase 1 occur rhythmic abrupt changes and fluctuations, which varying learning rates may cause. 
Nevertheless, all the losses decrease as the iteration  increases until they finally converge.
Since the convergence of the reconstruction loss is not apparent when all the losses are included in one picture, the reconstruction loss is enlarged in the blank space for further observation, as shown in the red dashed frame in Fig. 11. It is clear that the reconstruction loss gradually converges during continuous iterations.
In Phase 2, our goal is to train the decoder and the image classifier. Hence the total loss for Phase 2 is composed of reconstruction loss and image classification loss, as shown in Fig. 12. The image classification loss and the total loss exhibit downward trends on the whole and finally converge. Compared with Phase 1, the training process of this Phase fluctuates less and is more stable. The reconstruction loss is enlarged again in the blank place of this chart for further study. From the curve in the red dashed square frame, it is found that the reconstruction loss rises at the beginning, then decreases, and finally levels off. This phenomenon is probably due to the calculation method for the reconstruction loss. In Phase 1, the reconstruction loss is calculated based on the input image and the reconstructed image, which is obtained by decoding the latent feature encoded from the input image. Different from Phase 1, Phase 2 firstly samples from the latent variables' distribution trained in Phase 1 and then decodes the sampling point to get the synthesized image. Finally, the loss is calculated between the synthesized image and the real sample, which is randomly chosen. Hence, the generated images in Phase 2 are not the real reconstructed images for the input images but the generations from the latent features. Furthermore, the random sampling points have more possibilities, leading to the corresponding changes in generated samples.  Therefore, the reconstruction loss exceeds a relearning process instead of continuing to converge based on the previous loss.
In Phase 3, the MEDA\_LUDE aims at training the encoder and the latent feature classifier, and the total loss includes the L-GM loss and the latent feature classification loss. We can conclude from Fig. 13 that all the losses decrease stably until final convergence. 
To sum up, the training process in Phases 1-3 fully showcases the stability and the robustness of the MEDA\_LUDE model in high-dimensional images.

\begin{figure}[htbp]
  \centering 
  \includegraphics[width=.9\linewidth]{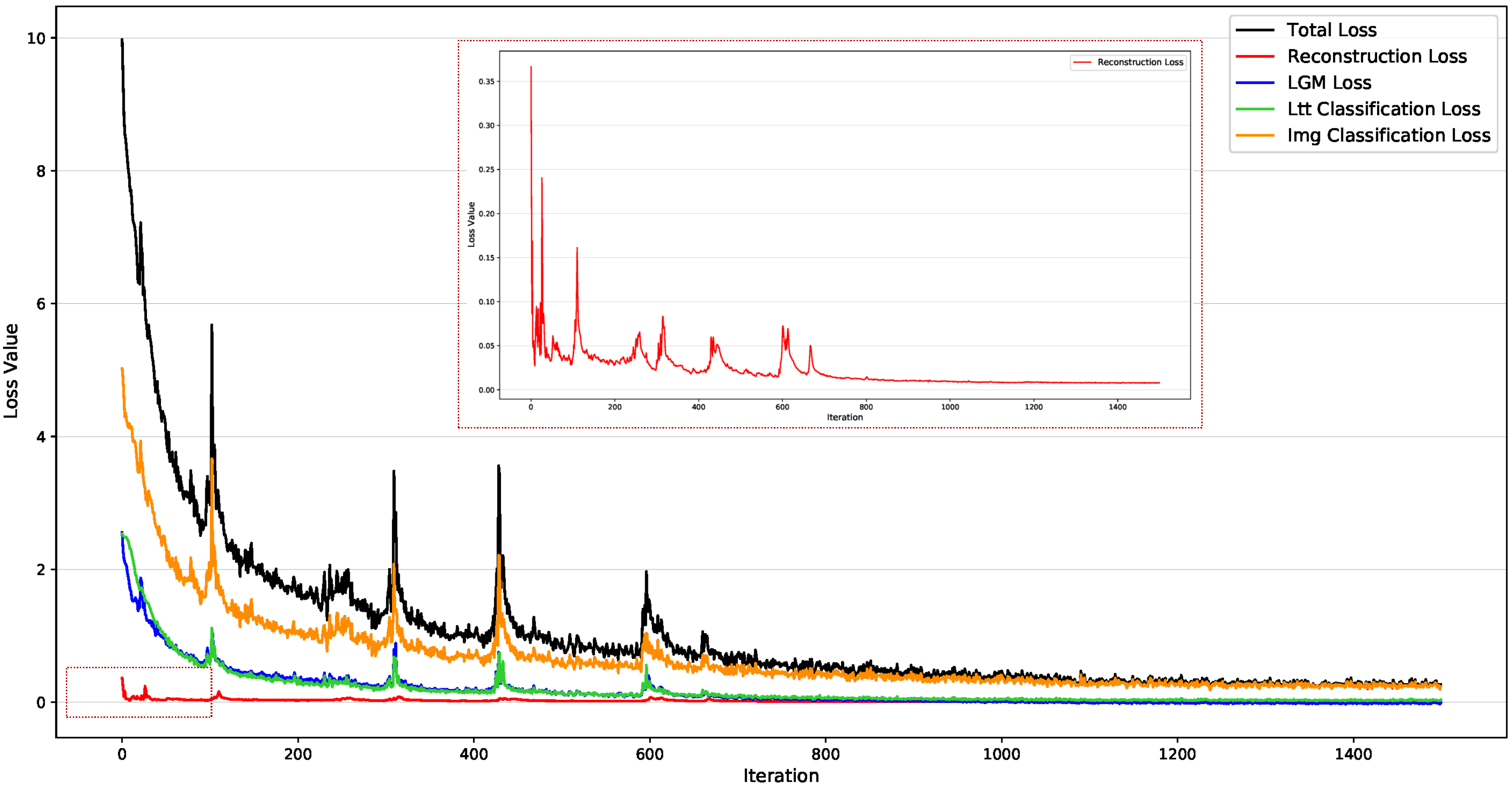}\\
  \caption{Losses in Phase 1 for DHU-FD dataset}\label{Fig. 1}
  \vspace{-0.2cm}
\end{figure}

\begin{figure}[htbp]
  \centering 
  \includegraphics[width=.9\linewidth]{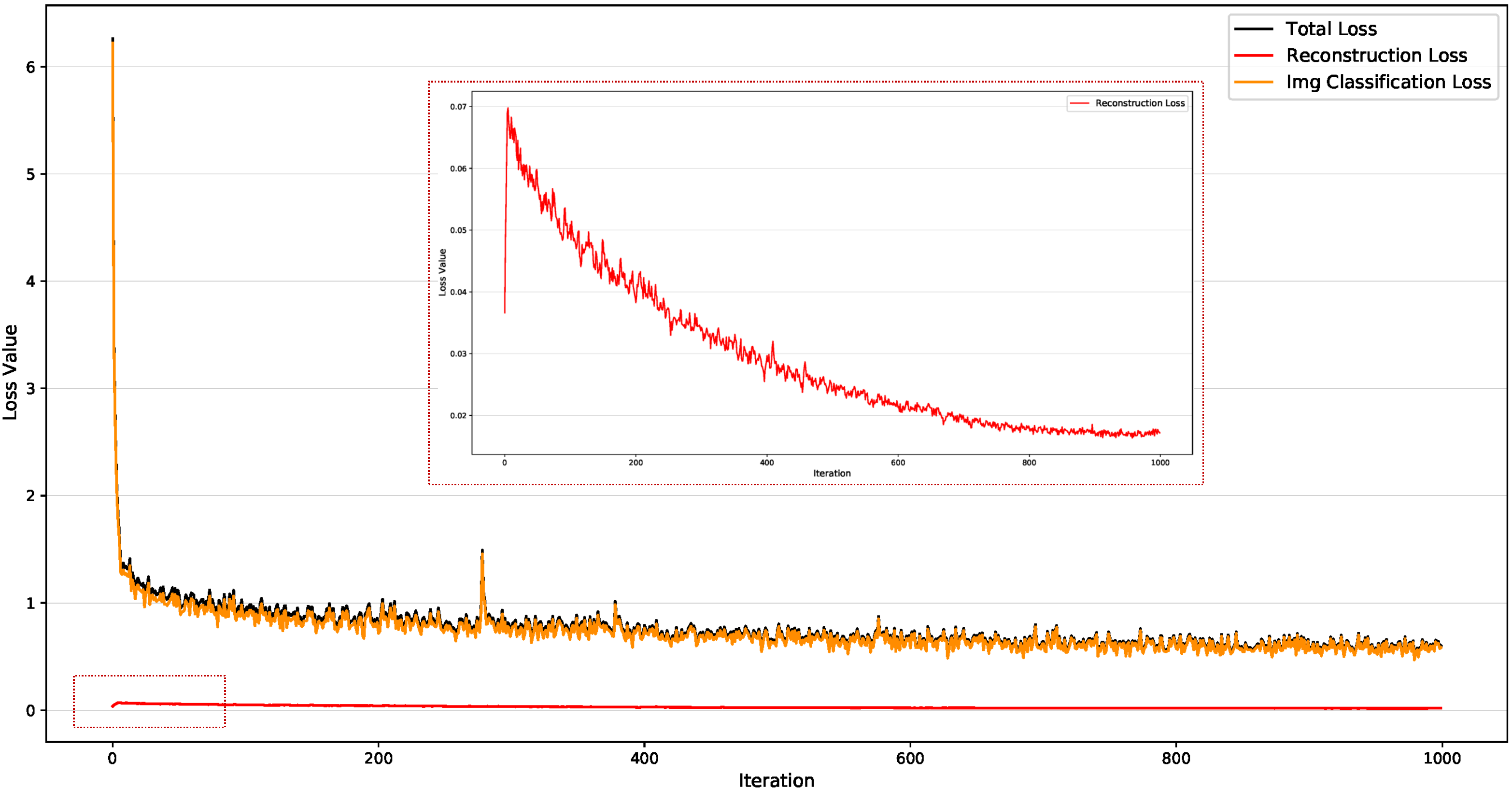}\\
  \caption{Losses in Phase 2 for DHU-FD dataset}\label{Fig. 1}
  \vspace{-0.2cm}
\end{figure}

\begin{figure}[htbp]
  \centering 
  \includegraphics[width=.9\linewidth]{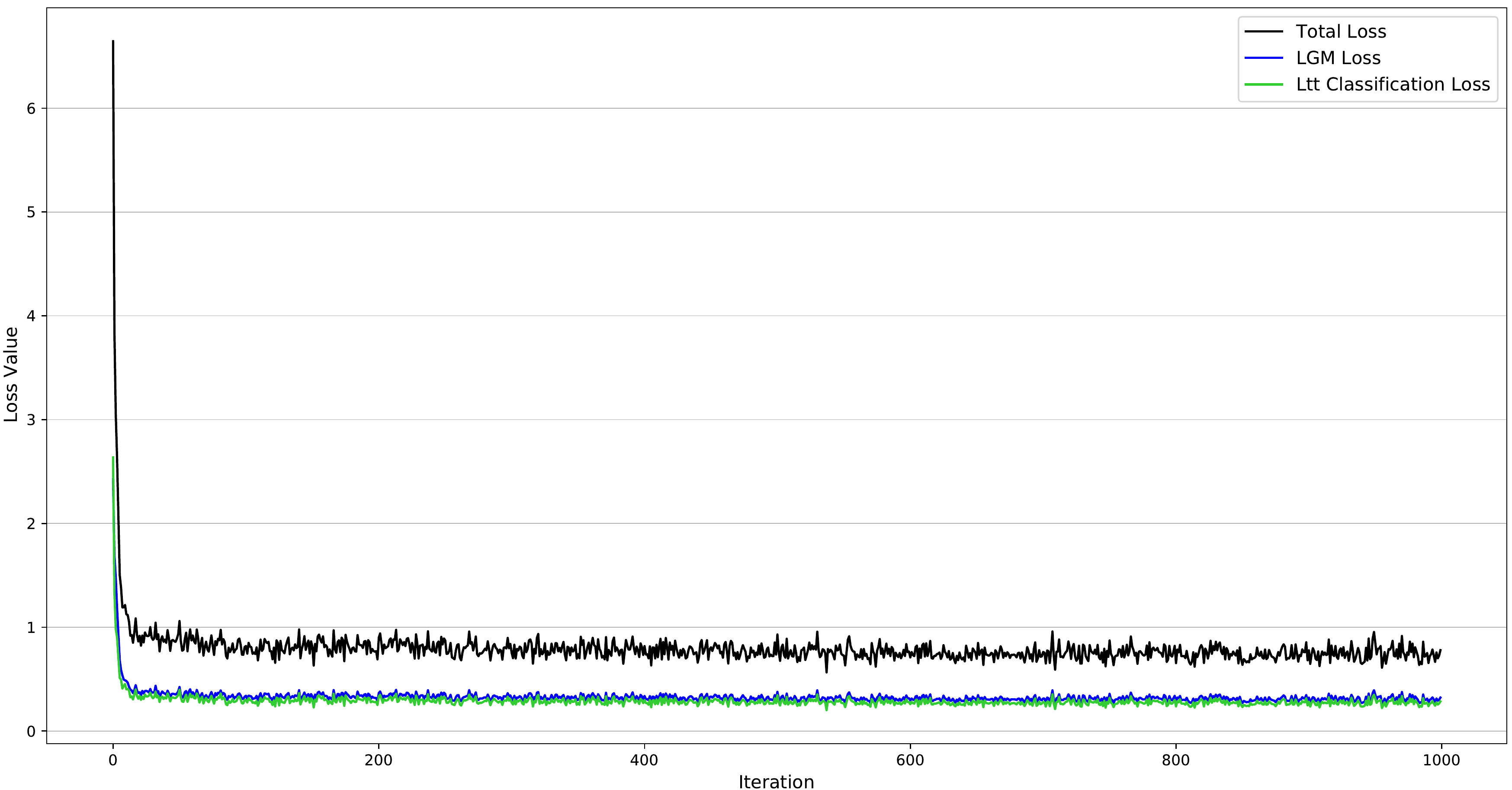}\\
  \caption{Losses in Phase 3 for DHU-FD dataset}\label{Fig. 1}
  \vspace{-0.2cm}
\end{figure}

\begin{table*}[!h]\footnotesize
\renewcommand\arraystretch{1}
\centering
\caption{Classification performance of the comparative methods for DHU-FD}
\begin{tabular}{p{1.2cm} p{0.6cm}  p{1cm}p{1cm} p{1cm} p{1cm} p{1cm} p{1.2cm} p{1.2cm}}
  \hline
  \hline
   Evaluation & Seed & RMR& ROS  & SMOTE & ADASYN  & CVAE & CVAE\_& \textbf{MEDA\_}\\ 
   Critieria&  &   & &  &   &  & SeTred&\textbf{LUDE}\\
  \Xhline{0.8pt}
   Accuracy &1&63.50&66.00&70.00&69.00&70.00&71.00&\textbf{74.00}\\
  (\%)&3&63.50&70.50&71.00&73.00&73.50&74.00&\textbf{76.00}\\
  \hline
   Precision&1&69.38&75.56&69.36&67.00&78.02&\textbf{79.44}&79.11\\
    (\%)&3&69.38&75.98&71.71&75.11&74.29&75.46&\textbf{79.09}\\
  \hline
   Recall&1&63.50&66.00&70.00&69.00&70.00&71.00&\textbf{74.00}\\
    (\%)&3&63.50&70.50&71.00&73.00&73.50&74.00&\textbf{76.00}\\
  \hline
      Specificity &1&95.94&96.22&96.67&96.56&96.67&96.78&\textbf{97.11}\\   
     (\%)&3&95.94&96.72&96.78&97.00&97.06&97.11&\textbf{97.33}\\
  \hline
   F1&1&56.92&62.42&68.59&65.44&67.60&67.89&\textbf{74.28}\\
    (\%)&3&56.92&68.93&67.54&71.21&71.41&72.76&\textbf{73.98}\\
  \hline
   GM &1&70.63&75.69&80.93&78.21&79.83&79.21&\textbf{84.13}\\ 
    (\%)&3&70.63&81.03&77.34&79.15&79.49&82.21&\textbf{83.31}\\
  \hline
  AUC &1&0.9444&0.9032&0.9439&0.9375&0.9538&0.9560&\textbf{0.9603}\\
  &3&0.9444&\textbf{0.9625}&0.9421&0.9270&0.9352&0.9391&0.9371\\
  \hline
  \hline
\end{tabular}
\end{table*}

\begin{table*}[!h]\footnotesize
\renewcommand\arraystretch{1}
\centering
\caption{Classification performance of the comparative methods for ALYUN-FD}
\begin{tabular}{p{1.2cm} p{0.6cm}  p{1cm}p{1cm} p{1cm} p{1cm} p{1cm} p{1.2cm} p{1.2cm}}
  \hline
  \hline
   Evaluation & Seed & RMR& ROS  & SMOTE & ADASYN  & CVAE & CVAE\_& \textbf{MEDA\_}\\ 
   Critieria&  &   & &  &   &  & SeTred&\textbf{LUDE}\\
  \Xhline{0.8pt}
   Accuracy &0&56.72&59.06&59.83&60.11&60.28&61.00&\textbf{62.06}\\
  (\%)&1&58.72&58.17&58.50&58.56&58.72&60.00&\textbf{60.94}\\
  \hline
   Precision&0&66.34&66.35&65.60&64.28&65.47&66.91&\textbf{66.94}\\
    (\%)&1&64.83&64.21&65.82&62.79&65.21&65.20&\textbf{66.03}\\
  \hline
   Recall &0&56.72&59.06&59.83&60.11&60.28&61.00&\textbf{62.06}\\
    (\%)&1&58.72&58.17&58.50&58.56&58.72&60.00&\textbf{60.94}\\
  \hline
      Specificity &0&91.34&91.81&91.97&92.02&92.06&92.20&\textbf{92.41}\\   
     (\%)&1&91.74&91.63&91.70&91.71&91.74&92.00&\textbf{92.19}\\
  \hline
   F1&0&51.72&56.19&58.34&57.69&57.61&58.28&\textbf{59.54}\\
    (\%)&1&54.56&55.63&56.08&55.56&55.39&57.39&\textbf{58.16}\\
  \hline
   GM &0&67.58&70.87&72.49&72.30&72.07&72.58&\textbf{73.49}\\ 
    (\%)&1&69.92&70.63&70.73&70.75&70.43&72.06&\textbf{72.59}\\
  \hline
  AUC &0&\textbf{0.8671}&0.8488&0.8418&0.8473&0.8474&0.8473&0.8569\\
        &1&\textbf{0.8714}&0.8440&0.8359&0.8380&0.8695&0.8389&0.8444\\
  \hline
  \hline
\end{tabular}
\end{table*}

The experimental results in DHU-FD dataset and ALIYUN-FD dataset are exhibited in TABLE \uppercase\expandafter{\romannumeral3} and TABLE \uppercase\expandafter{\romannumeral4}, respectively.
In TABLE \uppercase\expandafter{\romannumeral3}, the best accuracies under both the two random circumstances are achieved by the MEDA\_LUDE algorithm, and both of them are far superior to their respective suboptimal accuracies. Except for the precision of seed 1 and the AUC of seed 3, the MEDA\_LUDE algorithm outperforms all the other comparative methods in various criteria under two different imbalanced datasets randomly formed, which validates the effectiveness of the proposed method in handling the imbalanced problem for fabric defects from DHU.
For the ALIYUN-FD dataset, the proposed MEDA\_LUDE algorithm obtains the best results in all the indicators apart from the index AUC. When the random seed is 0, the most outstanding value for AUC is acquired by RMR, and our method's AUC is second only to RMR's. However, the accuracy of RMR is only $56.72\%$,  which is far smaller than ours, and the other indicators between RMR and the MEDA\_LUDE method still exist gaps.
When the random seed is set to 1, even though the MEDA\_LUDE doesn't get the best AUC, it still exposes the excellent classification performance in the other criteria.
Since the test set is balanced, according to the classification results presented above, it has proved that the MEDA\_LUDE algorithm can be applied successfully to the imbalanced classification task for fabric defects but still has room for further improvement.

\section{Conclusion}
In this paper, a new algorithm called MEDA\_LUDE is designed for trading off the quality and the diversity of generated samples in imbalanced image classification tasks. The proposed method combines deep neural networks and modified EDAs to optimize and evolve latent feature distributions jointly.  
Besides, the L-GM loss function is introduced to feature learning under a more complex assumption. The EDAs are improved to guide the searching space for better quality and diversity.
Classification performance on the two imbalanced datasets constructed from MNIST and CIFAR-10 datasets and the t-SNE visualizations indicate the efficacy of our method in generating samples with both the quality and the diversity.
Last, the success of applications in fabric defect classification shows the potential of the MEDA\_LUDE algorithm in solving imbalanced problems in the industry.

\appendices
\section*{Acknowledgment}
This work was supported in part by the Fundamental Research Funds for the Central Universities (2232021A-10, 2232021D-37), National Natural Science Foundation of China (61903078), Shanghai Sailing Program (22YF1401300), Natural Science Foundation of Shanghai (20ZR1400400, 21ZR1401700), 
and Chinese Ministry of Education Research Found on Intelligent Manufacturing (MCM20180703).

\ifCLASSOPTIONcaptionsoff
  \newpage
\fi

\bibliography{MEDA_LUDE}

\begin{thebibliography}{10}
\providecommand{\url}[1]{#1}
\csname url@samestyle\endcsname
\providecommand{\newblock}{\relax}
\providecommand{\bibinfo}[2]{#2}
\providecommand{\BIBentrySTDinterwordspacing}{\spaceskip=0pt\relax}
\providecommand{\BIBentryALTinterwordstretchfactor}{4}
\providecommand{\BIBentryALTinterwordspacing}{\spaceskip=\fontdimen2\font plus
\BIBentryALTinterwordstretchfactor\fontdimen3\font minus
  \fontdimen4\font\relax}
\providecommand{\BIBforeignlanguage}[2]{{%
\expandafter\ifx\csname l@#1\endcsname\relax
\typeout{** WARNING: IEEEtran.bst: No hyphenation pattern has been}%
\typeout{** loaded for the language `#1'. Using the pattern for}%
\typeout{** the default language instead.}%
\else
\language=\csname l@#1\endcsname
\fi
#2}}
\providecommand{\BIBdecl}{\relax}
\BIBdecl

\bibitem{zhang2016transfer}
X.~Zhang, Y.~Zhuang, W.~Wang, and W.~Pedrycz, ``Transfer boosting with
  synthetic instances for class imbalanced object recognition,'' \emph{IEEE
  Transactions on Cybernetics}, vol.~48, no.~1, pp. 357--370, 2016.

\bibitem{8528854}
Z.~Zhu, Z.~Wang, D.~Li, Y.~Zhu, and W.~Du, ``Geometric structural ensemble
  learning for imbalanced problems,'' \emph{IEEE Transactions on Cybernetics},
  vol.~50, no.~4, pp. 1617--1629, 2020.

\bibitem{rosales2022handling}
A.~Rosales-P{\'e}rez, S.~Garc{\'\i}a, and F.~Herrera, ``Handling imbalanced
  classification problems with support vector machines via evolutionary bilevel
  optimization,'' \emph{IEEE Transactions on Cybernetics}, 2022,
  doi:10.1109/TCYB.2022.3163974.

\bibitem{liu2007generative}
A.~Liu, J.~Ghosh, and C.~E. Martin, ``Generative oversampling for mining
  imbalanced datasets.'' in \emph{DMIN}, 2007, pp. 66--72.

\bibitem{chawla2002smote}
N.~V. Chawla, K.~W. Bowyer, L.~O. Hall, and W.~P. Kegelmeyer, ``Smote:
  synthetic minority over-sampling technique,'' \emph{Journal of Artificial
  Intelligence Research}, vol.~16, pp. 321--357, 2002.

\bibitem{han2005borderline}
H.~Han, W.-Y. Wang, and B.-H. Mao, ``Borderline-smote: a new over-sampling
  method in imbalanced data sets learning,'' in \emph{International Conference
  on Intelligent Computing}.\hskip 1em plus 0.5em minus 0.4em\relax Springer,
  2005, pp. 878--887.

\bibitem{he2008adasyn}
H.~He, Y.~Bai, E.~A. Garcia, and S.~Li, ``Adasyn: Adaptive synthetic sampling
  approach for imbalanced learning,'' in \emph{2008 IEEE International Joint
  Conference on Neural Networks (IEEE World Congress on Computational
  Intelligence)}.\hskip 1em plus 0.5em minus 0.4em\relax IEEE, 2008, pp.
  1322--1328.

\bibitem{barua2012mwmote}
S.~Barua, M.~M. Islam, X.~Yao, and K.~Murase, ``Mwmote--majority weighted
  minority oversampling technique for imbalanced data set learning,''
  \emph{IEEE Transactions on Knowledge and Data Engineering}, vol.~26, no.~2,
  pp. 405--425, 2012.

\bibitem{9216561}
J.~Chen, L.~Du, and L.~Liao, ``Discriminative mixture variational autoencoder
  for semisupervised classification,'' \emph{IEEE Transactions on Cybernetics},
  pp. 1--15, 2020.

\bibitem{wan2017variational}
Z.~Wan, Y.~Zhang, and H.~He, ``Variational autoencoder based synthetic data
  generation for imbalanced learning,'' in \emph{2017 IEEE Symposium Series on
  Computational Intelligence (SSCI)}.\hskip 1em plus 0.5em minus 0.4em\relax
  IEEE, 2017, pp. 1--7.

\bibitem{dai2019generative}
W.~Dai, K.~Ng, K.~Severson, W.~Huang, F.~Anderson, and C.~Stultz, ``Generative
  oversampling with a contrastive variational autoencoder,'' in \emph{2019 IEEE
  International Conference on Data Mining (ICDM)}.\hskip 1em plus 0.5em minus
  0.4em\relax IEEE, 2019, pp. 101--109.

\bibitem{lim2018molecular}
J.~Lim, S.~Ryu, J.~W. Kim, and W.~Y. Kim, ``Molecular generative model based on
  conditional variational autoencoder for de novo molecular design,''
  \emph{Journal of Cheminformatics}, vol.~10, no.~1, pp. 1--9, 2018.

\bibitem{yang2020generative}
Y.~Yang, C.~Malaviya, J.~Fernandez, S.~Swayamdipta, R.~L. Bras, J.-P. Wang,
  C.~Bhagavatula, Y.~Choi, and D.~Downey, ``Generative data augmentation for
  commonsense reasoning,'' \emph{arXiv preprint arXiv:2004.11546}, 2020.

\bibitem{9387399}
S.~Du, J.~Hong, Y.~Wang, and Y.~Qi, ``A high-quality multicategory sar images
  generation method with multiconstraint gan for atr,'' \emph{IEEE Geoscience
  and Remote Sensing Letters}, vol.~19, pp. 1--5, 2022.

\bibitem{yang2021data}
C.~Yang, Y.~Shen, Y.~Xu, and B.~Zhou, ``Data-efficient instance generation from
  instance discrimination,'' \emph{arXiv preprint arXiv:2106.04566}, 2021.

\bibitem{costa2020exploring}
V.~Costa, N.~Louren{\c{c}}o, J.~Correia, and P.~Machado, ``Exploring the
  evolution of gans through quality diversity,'' in \emph{Proceedings of the
  2020 Genetic and Evolutionary Computation Conference}, 2020, pp. 297--305.

\bibitem{alihosseini2019jointly}
D.~Alihosseini, E.~Montahaei, and M.~S. Baghshah, ``Jointly measuring diversity
  and quality in text generation models,'' in \emph{Proceedings of the Workshop
  on Methods for Optimizing and Evaluating Neural Language Generation}, 2019,
  pp. 90--98.

\bibitem{li2020relation}
J.~Li, Y.~Lan, J.~Guo, and X.~Cheng, ``On the relation between
  quality-diversity evaluation and distribution-fitting goal in text
  generation,'' in \emph{International Conference on Machine Learning}.\hskip
  1em plus 0.5em minus 0.4em\relax PMLR, 2020, pp. 5905--5915.

\bibitem{bontrager2018deepmasterprints}
P.~Bontrager, A.~Roy, J.~Togelius, N.~Memon, and A.~Ross, ``Deepmasterprints:
  Generating masterprints for dictionary attacks via latent variable
  evolution,'' in \emph{2018 IEEE 9th International Conference on Biometrics
  Theory, Applications and Systems (BTAS)}.\hskip 1em plus 0.5em minus
  0.4em\relax IEEE, 2018, pp. 1--9.

\bibitem{volz2018evolving}
V.~Volz, J.~Schrum, J.~Liu, S.~M. Lucas, A.~Smith, and S.~Risi, ``Evolving
  mario levels in the latent space of a deep convolutional generative
  adversarial network,'' in \emph{Proceedings of the Genetic and Evolutionary
  Computation Conference}, 2018, pp. 221--228.

\bibitem{giacomello2019searching}
E.~Giacomello, P.~L. Lanzi, and D.~Loiacono, ``Searching the latent space of a
  generative adversarial network to generate doom levels,'' in \emph{2019 IEEE
  Conference on Games (CoG)}.\hskip 1em plus 0.5em minus 0.4em\relax IEEE,
  2019, pp. 1--8.

\bibitem{thakkar2019autoencoder}
S.~Thakkar, C.~Cao, L.~Wang, T.~J. Choi, and J.~Togelius, ``Autoencoder and
  evolutionary algorithm for level generation in lode runner,'' in \emph{2019
  IEEE Conference on Games (CoG)}.\hskip 1em plus 0.5em minus 0.4em\relax IEEE,
  2019, pp. 1--4.

\bibitem{zhou2007survey}
S.~Zhou and Z.~Sun, ``A survey on estimation of distribution algorithms,''
  \emph{Acta Automatica Sinica}, vol.~33, no.~2, p. 113, 2007.

\bibitem{yang2016multimodal}
Q.~Yang, W.-N. Chen, Y.~Li, C.~P. Chen, X.-M. Xu, and J.~Zhang, ``Multimodal
  estimation of distribution algorithms,'' \emph{IEEE Transactions on
  Cybernetics}, vol.~47, no.~3, pp. 636--650, 2016.

\bibitem{liang2018enhancing}
Y.~Liang, Z.~Ren, X.~Yao, Z.~Feng, A.~Chen, and W.~Guo, ``Enhancing gaussian
  estimation of distribution algorithm by exploiting evolution direction with
  archive,'' \emph{IEEE Transactions on Cybernetics}, vol.~50, no.~1, pp.
  140--152, 2018.

\bibitem{xue2017Swarm}
X.~et~al., ``Swarm intelligence optimization based on estimation of
  distribution algorithm,'' \emph{Computer And Modernization}, vol. 000, no.
  001, pp. 17--22, 2017.

\bibitem{wan2018rethinking}
W.~Wan, Y.~Zhong, T.~Li, and J.~Chen, ``Rethinking feature distribution for
  loss functions in image classification,'' in \emph{Proceedings of the IEEE
  Conference on Computer Vision and Pattern Recognition}, 2018, pp. 9117--9126.

\bibitem{weng2011secure}
L.~Weng and B.~Preneel, ``A secure perceptual hash algorithm for image content
  authentication,'' in \emph{IFIP International Conference on Communications
  and Multimedia Security}.\hskip 1em plus 0.5em minus 0.4em\relax Springer,
  2011, pp. 108--121.

\bibitem{kang2016noise}
Q.~Kang, X.~Chen, S.~Li, and M.~Zhou, ``A noise-filtered under-sampling scheme
  for imbalanced classification,'' \emph{IEEE Transactions on Cybernetics},
  vol.~47, no.~12, pp. 4263--4274, 2016.

\bibitem{van2008visualizing}
L.~Van~der Maaten and G.~Hinton, ``Visualizing data using t-sne.''
  \emph{Journal of Machine Learning Research}, vol.~9, no.~11, 2008.

\end{thebibliography}

\end{document}